%% file: main.tex
\documentclass{article} %
\usepackage{iclr2027_conference,times}

\input{math_commands.tex}

\input{generated/results_macros.tex}

\usepackage{hyperref}
\usepackage{url}
\newcommand{\doi}[1]{doi: \href{https://doi.org/#1}{\nolinkurl{#1}}}
\usepackage{graphicx}
\usepackage{booktabs}
\usepackage{amsmath,amssymb}
\usepackage{xcolor}
\usepackage{subcaption}
\usepackage{microtype}
\usepackage{enumitem}

\graphicspath{{figures/}}

\newcommand{\model}{Llama\mbox{-}3.1\mbox{-}70B}
\newcommand{\nheads}{5{,}120}

\title{Which Attention Heads are like the Human Head? Not the Ones that Compute}

\author{%
\textbf{Christopher Pinier\textsuperscript{1,*}, Gustaw Opie\l{}ka\textsuperscript{1,*}, Hannes Rosenbusch\textsuperscript{1}} \\
\textbf{Taylor Webb\textsuperscript{2,+}, Michael D. Nunez\textsuperscript{1,+}, Claire E. Stevenson\textsuperscript{1,+}} \\[0.5em]
\normalfont\small \textsuperscript{1}Psychological Methods, University of Amsterdam, Amsterdam, The Netherlands \\
\normalfont\small \textsuperscript{2}Princeton Neuroscience Institute, Princeton University, Princeton, NJ, USA
}
\hypersetup{pdfauthor={Christopher Pinier, Gustaw Opiełka, Hannes Rosenbusch, Taylor Webb, Michael D. Nunez, Claire E. Stevenson}}

\iclrfinalcopy
\begin{document}

\maketitle
\lhead{}
\vspace{-0.2in}
{\small\raggedright
\noindent\hspace*{\tabcolsep}\textsuperscript{*}These authors contributed equally\\
\hspace*{\tabcolsep}\textsuperscript{+}Shared senior authorship\par}
\input{sections/00_abstract}

\begin{figure}[!h]
    \centering
    \includegraphics[width=0.95\linewidth]{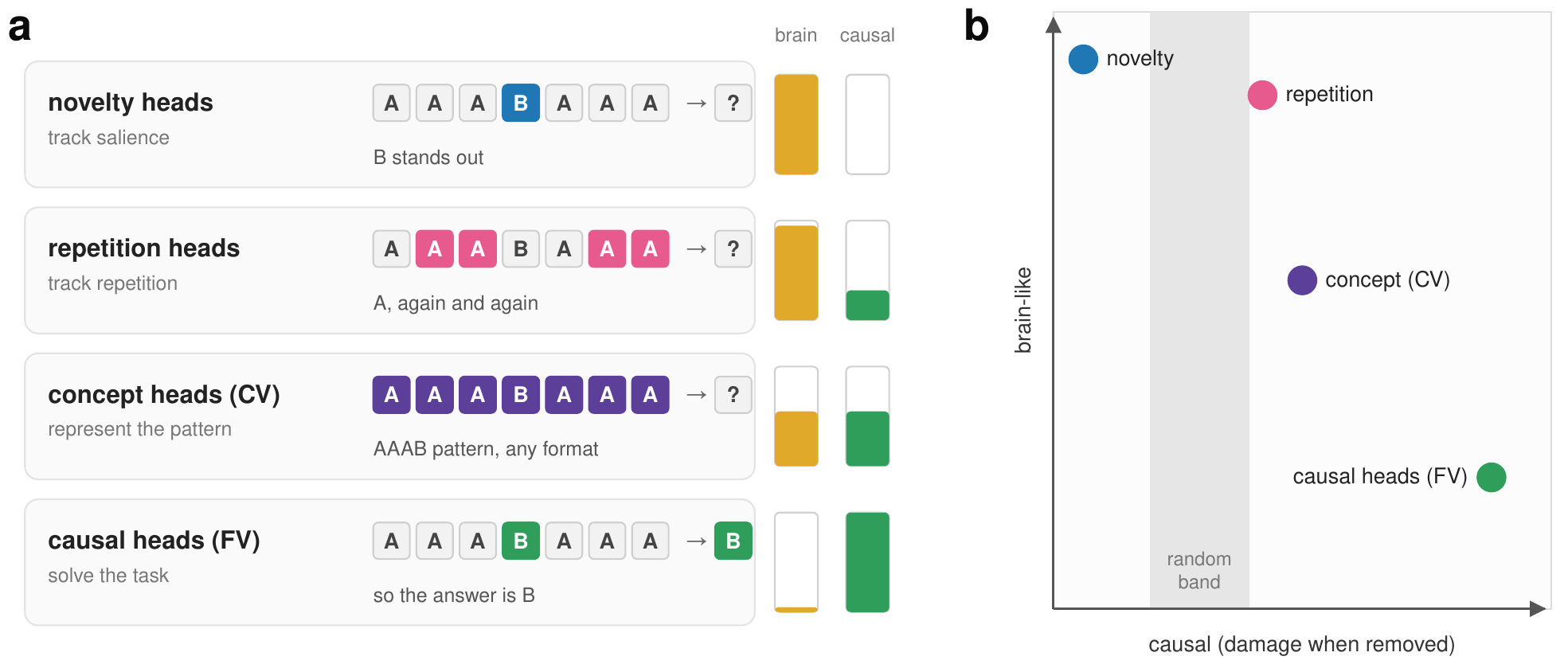}
    \caption{\textbf{Schematic description of the findings.}
    (a)~What each attention head population does with one item of the task, the pattern \texttt{AAABAAA} whose answer is B (in words, for example, ``hat hat hat risk hat hat hat'' with answer ``risk''), beside its brain alignment and its causal contribution. Colored symbols mark what each population attends to or encodes.
    (b)~The same populations, schematically, in the plane of brain alignment against damage when the heads are removed; the gray band marks the damage from ablating random heads. Measured ablation curves are in Figure~\ref{fig:replication-brain-ablation}.}
    \label{fig:teaser}
\end{figure}

\input{sections/01_intro}

\input{sections/03_brain_causal}

\input{sections/04_profiles}

\input{sections/05_families}

\input{sections/06_gaze}
\input{sections/07_discussion}

\subsubsection*{Ethics statement}

The EEG and eye-tracking data analyzed here are drawn from the de-identified dataset described by \citet{pinier2025large}. This work reanalyzed this data and did not collect any additional human data. We release or share only de-identified data and derived representations.

\subsubsection*{AI use statement}

Generative AI tools were used for code assistance and for preliminary drafting and editing of prose. They were not used to generate human-participant data, to make analytic decisions without author review, or to fabricate experimental results, citations, or figures.

\newpage
\bibliography{references}
\bibliographystyle{iclr2027_conference}

\newpage
\appendix
\input{sections/08_appendix_A}

\end{document}

%% file: math_commands.tex
\usepackage{amsmath,amsfonts,bm}

\def\eqref#1{equation~\ref{#1}}

\def\1{\bm{1}}

\DeclareMathAlphabet{\mathsfit}{\encodingdefault}{\sfdefault}{m}{sl}
\SetMathAlphabet{\mathsfit}{bold}{\encodingdefault}{\sfdefault}{bx}{n}

%% file: generated/results_macros.tex
\providecommand{\paperresult}[1]{\ifcsname paperresult@#1\endcsname\csname paperresult@#1\endcsname\else\PackageError{paperresults}{Unknown result ID: #1}{Check results_registry.yaml}\fi}
\expandafter\def\csname paperresult@clean.qwen2.5_14b\endcsname{52.5}
\expandafter\def\csname paperresult@clean.qwen2.5_14b_instruct\endcsname{75.0}
\expandafter\def\csname paperresult@clean.qwen2.5_32b\endcsname{74.5}
\expandafter\def\csname paperresult@clean.qwen2.5_32b_instruct\endcsname{77.5}
\expandafter\def\csname paperresult@clean.qwen2.5_3b\endcsname{29.5}
\expandafter\def\csname paperresult@clean.qwen2.5_3b_instruct\endcsname{25.0}
\expandafter\def\csname paperresult@clean.qwen2.5_72b\endcsname{54.0}
\expandafter\def\csname paperresult@clean.qwen2.5_72b_instruct\endcsname{87.5}
\expandafter\def\csname paperresult@clean.deepseek_r1_distill_llama_70b\endcsname{63.0}
\expandafter\def\csname paperresult@clean.llama_3.1_70b\endcsname{70.5}
\expandafter\def\csname paperresult@clean.llama_3.1_70b_instruct\endcsname{77.5}
\expandafter\def\csname paperresult@clean.llama_3.1_8b\endcsname{62.5}
\expandafter\def\csname paperresult@clean.llama_3.1_8b_instruct\endcsname{59.5}
\expandafter\def\csname paperresult@clean.llama_3.2_3b\endcsname{51.0}
\expandafter\def\csname paperresult@clean.llama_3.2_3b_instruct\endcsname{39.5}
\expandafter\def\csname paperresult@clean.llama_3.3_70b_instruct\endcsname{82.5}
\expandafter\def\csname paperresult@clean.phi_4\endcsname{82.0}
\expandafter\def\csname paperresult@cohort.models\endcsname{17}
\expandafter\def\csname paperresult@cohort.additional_models\endcsname{16}
\expandafter\def\csname paperresult@inference.llama8.open_ended\endcsname{91.5}
\expandafter\def\csname paperresult@inference.llama8.multiple_choice\endcsname{62.5}
\expandafter\def\csname paperresult@inference.llama70.open_ended\endcsname{90.0}
\expandafter\def\csname paperresult@inference.llama70.multiple_choice\endcsname{68.5}
\expandafter\def\csname paperresult@inference.llama33.words_eng.raw\endcsname{84.5}
\expandafter\def\csname paperresult@inference.llama33.words_eng.chat\endcsname{91.0}
\expandafter\def\csname paperresult@inference.llama33.symbols.raw\endcsname{33.5}
\expandafter\def\csname paperresult@inference.llama33.symbols.chat\endcsname{80.5}
\expandafter\def\csname paperresult@inference.symbols.min\endcsname{18.5}
\expandafter\def\csname paperresult@inference.symbols.max\endcsname{33.0}
\expandafter\def\csname paperresult@brain.concept.llama.all.min\endcsname{.100}
\expandafter\def\csname paperresult@brain.concept.llama.all.max\endcsname{.325}
\expandafter\def\csname paperresult@brain.concept.qwen.base.min\endcsname{.022}
\expandafter\def\csname paperresult@brain.concept.qwen.instruct.min\endcsname{-.205}
\expandafter\def\csname paperresult@brain.concept.qwen.base.max\endcsname{.206}
\expandafter\def\csname paperresult@brain.concept.qwen.instruct.max\endcsname{-.037}
\expandafter\def\csname paperresult@brain.concept.deepseek\endcsname{.269}
\expandafter\def\csname paperresult@brain.concept.phi\endcsname{-.047}
\expandafter\def\csname paperresult@brain.ap.min\endcsname{-.017}
\expandafter\def\csname paperresult@brain.ap.max\endcsname{.077}
\expandafter\def\csname paperresult@brain.concept.mean\endcsname{.10}
\expandafter\def\csname paperresult@brain.concept.min\endcsname{-.20}
\expandafter\def\csname paperresult@brain.concept.max\endcsname{.33}
\expandafter\def\csname paperresult@brain.ap.mean\endcsname{.017}
\expandafter\def\csname paperresult@carriers.all.repetition\endcsname{17.5}
\expandafter\def\csname paperresult@carriers.all.novelty\endcsname{46.7}
\expandafter\def\csname paperresult@carriers.top20.repetition\endcsname{22.6}
\expandafter\def\csname paperresult@carriers.top20.novelty\endcsname{44.4}
\expandafter\def\csname paperresult@carriers.top20.neither\endcsname{32.9}
\expandafter\def\csname paperresult@clusters.top20.min\endcsname{3}
\expandafter\def\csname paperresult@clusters.top20.max\endcsname{13}
\expandafter\def\csname paperresult@clusters.matches.novelty\endcsname{17}
\expandafter\def\csname paperresult@clusters.matches.repetition\endcsname{13}
\expandafter\def\csname paperresult@overlap.repetition\endcsname{.745}
\expandafter\def\csname paperresult@overlap.exact.repetition\endcsname{4}
\expandafter\def\csname paperresult@overlap.novelty\endcsname{.742}
\expandafter\def\csname paperresult@overlap.exact.novelty\endcsname{4}
\expandafter\def\csname paperresult@gaze.rdm.novelty\endcsname{.494}
\expandafter\def\csname paperresult@gaze.rdm.repetition\endcsname{.308}
\expandafter\def\csname paperresult@gaze.top1.novelty.relative.min\endcsname{.318}
\expandafter\def\csname paperresult@gaze.top1.novelty.relative.max\endcsname{.421}
\expandafter\def\csname paperresult@gaze.top1.repetition.relative.min\endcsname{-.335}
\expandafter\def\csname paperresult@gaze.top1.repetition.relative.max\endcsname{-.206}
\expandafter\def\csname paperresult@gaze.top1.cv.relative.min\endcsname{-.222}
\expandafter\def\csname paperresult@gaze.top1.cv.relative.max\endcsname{.438}
\expandafter\def\csname paperresult@gaze.top1.fv.relative.min\endcsname{-.127}
\expandafter\def\csname paperresult@gaze.top1.fv.relative.max\endcsname{.510}
\expandafter\def\csname paperresult@gaze.top1.novelty.direct\endcsname{-.051}
\expandafter\def\csname paperresult@gaze.top1.novelty.relative\endcsname{.373}
\expandafter\def\csname paperresult@gaze.top1.repetition.direct\endcsname{-.094}
\expandafter\def\csname paperresult@gaze.top1.repetition.relative\endcsname{-.285}
\expandafter\def\csname paperresult@gaze.top1.cv.relative\endcsname{.054}
\expandafter\def\csname paperresult@gaze.top1.fv.relative\endcsname{.180}
\expandafter\def\csname paperresult@gaze.top1.top.relative\endcsname{.180}
\expandafter\def\csname paperresult@gaze.top0.5.novelty.relative\endcsname{.366}
\expandafter\def\csname paperresult@gaze.top0.5.repetition.relative\endcsname{-.287}
\expandafter\def\csname paperresult@ablation.peak.fv.loss\endcsname{42.6}
\expandafter\def\csname paperresult@ablation.peak.fv.percent\endcsname{12.5}
\expandafter\def\csname paperresult@ablation.peak.cv.loss\endcsname{20.7}
\expandafter\def\csname paperresult@ablation.peak.cv.percent\endcsname{19.5}
\expandafter\def\csname paperresult@ablation.peak.brain.loss\endcsname{12.8}
\expandafter\def\csname paperresult@ablation.peak.brain.percent\endcsname{24.5}
\expandafter\def\csname paperresult@ablation.peak.repetition.loss\endcsname{11.2}
\expandafter\def\csname paperresult@ablation.peak.repetition.percent\endcsname{15.5}
\expandafter\def\csname paperresult@ablation.peak.novelty.loss\endcsname{0.3}
\expandafter\def\csname paperresult@exploration.ablation.activation_patching.option_chance_crossing.heads\endcsname{125}
\expandafter\def\csname paperresult@exploration.ablation.activation_patching.zero_crossing.heads\endcsname{800}
\expandafter\def\csname paperresult@exploration.ablation.novelty.at_800.accuracy\endcsname{63.5}
\expandafter\def\csname paperresult@exploration.ablation.novelty.at_800.clean_accuracy\endcsname{71.5}
\expandafter\def\csname paperresult@exploration.pythia.final.novelty32.ablated_accuracy\endcsname{39.5}
\expandafter\def\csname paperresult@exploration.pythia.final.novelty32.clean_accuracy\endcsname{42.5}
\expandafter\def\csname paperresult@exploration.pythia.final.novelty32.n_heads_ablated\endcsname{32}
\expandafter\def\csname paperresult@exploration.pythia.training.n_checkpoints\endcsname{19}
\expandafter\def\csname paperresult@exploration.scores.brain_activation_patching.correlation\endcsname{.03}
\expandafter\def\csname paperresult@exploration.scores.brain_concept.correlation\endcsname{.23}

%% file: sections/00_abstract.tex
\begin{abstract}
Brain–AI alignment is often interpreted as a sign that model and brain perform similar computations. Whether the aligned units are causally involved in model computation is rarely checked. On an abstract pattern-completion task (AAABAAA → B), we compare LLM attention-head representations with human EEG and test how ablating those heads affects task performance. Alignment and causation dissociate: brain-aligned heads contribute to performance, but their removal is substantially less disruptive than removal of heads selected via attribution patching. We compare two head sets that prior interpretability work defines without reference to the brain: concept vectors (CVs), which represent abstract patterns across formats, and function vectors (FVs), selected for their contribution to correct-answer prediction. Brain alignment shows little association with FV scores, while its association with CV scores varies across models. Among brain-aligned heads, we find recurring attention profiles: one emphasizes distinctive elements (novelty heads), the other repeating elements (repetition heads). The novelty family tracks salience and attends to the same elements that humans look at, yet its removal is less damaging than random ablation on average. Repetition heads contribute modestly to performance and are associated with abstract-pattern representation (CVs). Across 17 models spanning 3B–72B parameters, FV-ranked removal is substantially more disruptive than brain-ranked removal. Brain alignment thus captures how the model reads the stimulus, and only faintly captures how it represents the pattern and solves the task.
\end{abstract}

%% file: sections/01_intro.tex
\section{Introduction}
\label{sec:intro}

When a unit inside an artificial neural network predicts a neural signal recorded from a person performing the same task, the result is reported as \emph{alignment} and often read as evidence that the two systems compute something in common \citep{yamins2014performance,schrimpf2018brainscore,schrimpf2021neural,caucheteux2022brains,goldstein2022shared}. However, that reading has been challenged on logical and empirical grounds \citep{guest2023logical,bowers2023deep,antonello2024predictive}. The inference aspect has been tested indirectly: training a model to be brain-misaligned degrades its downstream performance \citep{merlin2026lose}, and units picked out by a neuroscience-style localizer are both causal and brain-aligned \citep{alkhamissi2025llm}. What has not been done is the direct test: select units by their alignment with a neural signal recorded on a task, remove them, and measure the model's performance on that same task. In large language models (LLMs), attention heads provide a tractable unit for this test: their representations can be compared with neural recordings, and their contribution to task performance can be assessed by removing or patching their outputs. They also bring a prior that neural data alone cannot supply: mechanistic interpretability has already identified head classes with known roles, such as induction and copy heads \citep{olsson2022context} and the function-vector heads of in-context learning \citep{todd2024function}, so a brain-alignment score can be read against heads whose function is known rather than against anonymous units.

We ask whether attention heads whose representations resemble human brain activity are also those on which the model's task performance most depends. We investigate this question using the abstract-sequence-completion task of \citet{pinier2025large}: participants infer the missing eighth element from a sequence of icons arranged as one of eight distinct abstract patterns, such as $\texttt{AAABAAA}\rightarrow\texttt{B}$. Models receive text-based versions, with words or symbols replacing the icons (Figure~\ref{fig:task}). Using fixation-related potentials (FRPs), EEG signals time-locked to eye-fixation onset, \citet{pinier2025large} reported positive representational correlations between frontal activity and LLM layers selected for their encoding of abstract patterns. We extend this layer-level comparison to individual attention heads and test its relationship to performance.

In LLMs, a dissociation between conceptual encoding (concept vectors; CVs) and representations that causally support task performance (function vectors; FVs) has been highlighted by \citet{opielka2026causality}, building on the function-vector framework of \citet{todd2024function}. On verbal analogies, heads selected for format-invariant concept representation generalized across formats and languages, whereas heads selected by activation patching supported in-context performance despite less consistent representations across formats. This distinction motivates two complementary analyses of the sequence-compltetion task. First, we score heads for concept representation and contribution to performance, independently of human data, and compare these scores with brain alignment. Second, we select the most brain-aligned heads, characterize their attention profiles, and test the effects of removing them. These analyses ask whether brain resemblance is more closely associated with abstract-pattern representation, with heads that matter most for predicting the answer or neither.

Our findings are as follows:
\begin{enumerate}[leftmargin=*,itemsep=1pt]
    \item \textbf{Brain alignment does not track causal importance}
    (Section~\ref{sec:dissociation}). Brain scores are weakly associated with attribution patching effects, and across models, removing brain-ranked heads is far less damaging than removing FV-ranked heads, the most causally important set.

    \item \textbf{Brain-aligned heads reveal repetition and novelty profiles}
    (Section~\ref{sec:families}). Across models, two attention profiles consistently appear: one attends to repeated elements (the As in \texttt{AAABAAA}), the other to the distinctive element (the lone B in the same sequence).

    \item \textbf{Novelty heads contribute little to task performance}
    (Sections~\ref{sec:families} and~\ref{sec:pricing}). Novelty-ranked removal is less damaging than random removal on average. Additional tasks and training-stage analyses support this limited task contribution.

    \item \textbf{Repetition heads track abstract representations and contribute modestly to task performance}
    (Section~\ref{sec:pricing}).
    Repetition heads correlate with format-invariant concept representation (CVs) and co-occur with CV heads across layers. Removing them impairs performance more than random removal, yet leaves the CV representation intact.

    \item \textbf{Novelty-like attention mirrors human gaze shifts}
    (Section~\ref{sec:brain}). Across all tested models, novelty-like attention follows pattern-dependent changes in human gaze allocation, whereas repetition-like attention shows the opposite relationship. 
\end{enumerate}

\input{sections/02_task_models}

%% file: sections/02_task_models.tex
\begin{figure}[t]
    \centering
    \includegraphics[width=\linewidth]{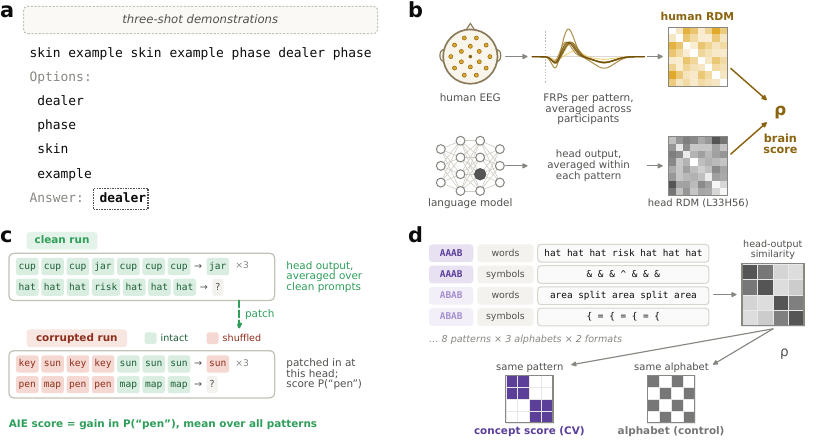}
    \caption{\textbf{The task prompt and the three head scores.} (a)~A query as presented to the model, preceded by three solved demonstrations in the same format (not shown here). The seven query symbols are based on one of eight abstract patterns (here, \texttt{ABABCDC → D}) and the model's next token after ``Answer:'' is scored. (b)~Brain score: Spearman correlation between a head's representational dissimilarity matrix (RDM), built from its output averaged within each pattern, and the human frontal-FRP RDM. (c)~Patching score: a head's output, averaged over clean prompts, is patched into a corrupted prompt from a different item. In the corrupted prompt, the elements of the first block (e.g., \texttt{AAAB} in \texttt{AAABAAA}) of every in-context demonstration and of the query are re-drawn at random from the item's symbols (red). This destroys the pattern while the remaining positions stay intact (green), and the demonstration answers are random. The gain in the probability of the clean answer (here ``pen'', which completes \texttt{AAABAAAB}), averaged over items from all eight patterns, is the head's average indirect effect (AIE); the heads with the highest AIE are the function-vector (FV) heads. (d)~Concept score: RSA of head outputs over 1,200 items that vary in pattern (8), alphabet (3; German, English, Symbols) and response format (2; open-ended, multiple choice), against a same-pattern design; a same-alphabet design serves as a control.}
    \label{fig:scores}
    \label{fig:task}
\end{figure}

\paragraph{Current study.}
\label{sec:setup}

The abstract-sequence experiment of \citet{pinier2025large} combined EEG and eye tracking during 400 unique trials. Our brain-alignment target is an $8\times8$ representational dissimilarity matrix (RDM) of group-average frontal FRPs: rows and columns index the eight abstract patterns, and entries measure dissimilarity between their FRP waveforms. We additionally use human gaze data as a separate comparison of attention profiles. Human data recording, preprocessing, and RDM construction are detailed in Appendix~\ref{app:human-study}.

We first investigated \model{} base \citep{grattafiori2024llama3} and then evaluated the generality of our findings in \paperresult{cohort.additional_models} additional models spanning 3B--72B parameters. Cross-model summaries include \model{}, giving \paperresult{cohort.models} models in total. The cohort covers base and instruction-tuned Llama and Qwen variants, plus Phi-4 and DeepSeek-R1-Distill-Llama-70B. We present the initial and cross-model findings together, using the same human FRP target and noting protocol differences where relevant. Throughout the experiment, we use a text-based adaptation of the human task and raw prompts (i.e., no chat template) built with three solved examples.

The initial \model{} evaluation achieved 70.5\% accuracy. Across the cohort, larger models generally performed better within Llama and instruction-tuned Qwen, but size alone did not determine success: Qwen2.5 base scored \paperresult{clean.qwen2.5_32b}\% at 32B versus \paperresult{clean.qwen2.5_72b}\% at 72B, while Phi-4 reached \paperresult{clean.phi_4}\% at 14B. Instruction tuning improved Llama-3.1-70B performance but reduced it for Llama-3.1-8B under the same raw-prompt protocol. Thus, performance depends on model family and training variant as well as scale. These immediate-answer scores do not measure each model's best achievable performance or the benefit of extended reasoning.

%% file: sections/03_brain_causal.tex
\begin{figure}[t]
    \centering
    \includegraphics[width=\linewidth]{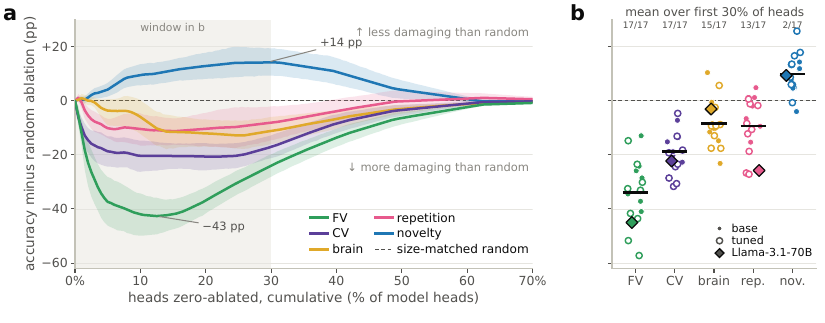}
    \caption{\textbf{Cumulative ablation relative to random removal, across 17 models.} Heads are zero-ablated cumulatively in each family's ranking order; values are accuracy minus the mean of five size-matched random ablations, in percentage points (pp), so negative values mean greater impairment than random. (a)~Mean over models, with head counts interpolated on a percentage-of-heads axis; bands are model-bootstrap 95\% intervals. FV-ranked ablations are the most damaging ($-43$~pp at 12.5\% of heads), whereas removing novelty-ranked heads is less damaging than random removal ($14$~pp at 30\%). (b)~The same difference for each model, averaged over the first 30\% of heads removed (shaded in a). Filled points are base models, open points instruction-tuned or distilled, and the diamond is \model{}; bars are means over models, and counts give the models below random controls.}
    \label{fig:replication-brain-ablation}
\end{figure}

\section{Brain alignment versus causal importance}
\label{sec:dissociation}

Does similarity to human brain activity identify the heads on which task performance depends, or those that represent the task's abstract patterns? We address this question first by comparing brain alignment with concept and patching scores across heads, then by testing the effects of removing heads selected by each criterion.

\paragraph{Three head scores.}
We scored every attention head by three criteria, illustrated in Figure~\ref{fig:scores}b--d. The \emph{brain score} (Figure~\ref{fig:scores}b) measures how closely a head's pattern-level output geometry resembles the frontal-FRP RDM. The \emph{patching score} (Figure~\ref{fig:scores}c) estimates the average indirect effect on correct-answer prediction under activation patching, identifying candidate FV heads \citep{vig2020investigating,todd2024function}. The \emph{concept score} (Figure~\ref{fig:scores}d) measures correspondence with abstract pattern identity across presentation conditions (such as English vs. German words and multiple-choice vs. open-ended) using a procedure akin to that of \citet{opielka2026causality}. In the cohort replications we estimated the activation-patching score by attribution patching for efficiency, as the two closely agreed in effect magnitude (Pearson $r=.983$) and selected 95\% of the same top-20 heads in \model{} (Appendix~\ref{app:ap}).

\paragraph{Brain scores do not predict patching scores.}
We paired each head's brain score with its patching score and computed Spearman rank correlations across all heads of a model. The comparison asks whether heads ranked higher for brain alignment also rank higher for causal effect. In \model{}, the two were only weakly related ($\rho=\paperresult{exploration.scores.brain_activation_patching.correlation}$), and the same held in every model (mean $\rho=\paperresult{brain.ap.mean}$, range $[\paperresult{brain.ap.min}$, $\paperresult{brain.ap.max}]$; Figure~\ref{fig:brain-corr}). Within a model, heads with higher brain scores therefore do not have larger patching effects.

\paragraph{Brain alignment relates more to concept representation, but varies across models.}
Brain--concept correlations averaged $\rho=\paperresult{brain.concept.mean}$ across the \paperresult{cohort.models} models (range $[\paperresult{brain.concept.min}, \paperresult{brain.concept.max}]$), and were generally more positive and more variable than brain--patching correlations (Figure~\ref{fig:replication-brain-correlations}). The positive association observed in the exploratory \model{} analysis ($\rho=\paperresult{exploration.scores.brain_concept.correlation}$) extended to every Llama model in the replication cohort ($\rho\in[\paperresult{brain.concept.llama.all.min}, \paperresult{brain.concept.llama.all.max}]$) and to DeepSeek ($\rho=\paperresult{brain.concept.deepseek}$). Qwen showed a different pattern: associations were positive, though sometimes close to zero, in base models ($[\paperresult{brain.concept.qwen.base.min}, \paperresult{brain.concept.qwen.base.max}]$), but negative in all instruction-tuned variants ($[\paperresult{brain.concept.qwen.instruct.min}, \paperresult{brain.concept.qwen.instruct.max}]$). Phi-4 showed little association ($\rho=\paperresult{brain.concept.phi}$). Thus, brain alignment often accompanies abstract-pattern representation, especially in Llama, whereas its relationship with patching effects remains consistently small.

\paragraph{Brain-ranked ablation is the least damaging.}
We zero-ablated heads cumulatively in each score's order, from highest to lowest, and compared the accuracy trajectory with matched random ablations (Appendix~\ref{app:cvfv}; Appendix~\ref{app:pipeline-ablation}). In \model{}, on the 200 English multiple-choice trials, patching-ranked removal dropped accuracy from 71.5\% to below the 25\% four-option chance level at \paperresult{exploration.ablation.activation_patching.option_chance_crossing.heads} heads (2.44\% of all heads) and to zero at \paperresult{exploration.ablation.activation_patching.zero_crossing.heads} heads (15.63\% of all heads).
 Brain-ranked removal stayed close to the random trajectory.

The same ordering held in all 17 models (Figure~\ref{fig:replication-brain-ablation}), but there brain-ranked removal was not equivalent to random: averaged over the first 30\% of heads removed, it was more damaging than random in 15 of 17 models, against 17 of 17 for FV and CV. Peak accuracy loss beyond random ablation was \paperresult{ablation.peak.fv.loss}~pp at \paperresult{ablation.peak.fv.percent}\% removal for FV-ranked heads, \paperresult{ablation.peak.cv.loss}~pp at \paperresult{ablation.peak.cv.percent}\% for CV, and \paperresult{ablation.peak.brain.loss}~pp at \paperresult{ablation.peak.brain.percent}\% for brain (peaks of the mean trajectories). Brain-aligned heads therefore carried some causal load, but far less than FV or CV heads, and the brain-ranked curve reached its peak only after removing far more heads.

%% file: sections/04_profiles.tex
\section{Repetition- and novelty-like attention profiles}
\label{sec:families}

Do the heads most aligned with human FRPs share recognizable attention profiles, and do these profiles identify heads that matter for task performance? We first characterize their attention, then compare the resulting profiles and ablation effects across models.

\begin{figure}[t]
    \centering
    \includegraphics[width=\linewidth]{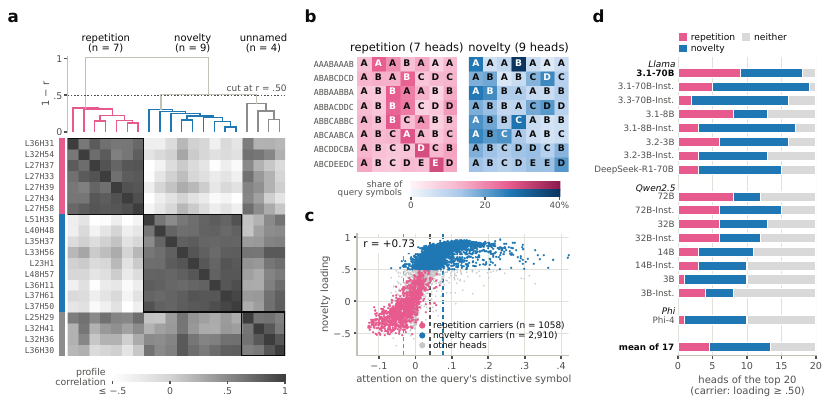}
    \caption{\textbf{Two attention profiles discovered among brain-aligned heads, and their recurrence across models.} (a)~The 20 highest brain-scoring heads in \model{}, clustered by their attention profiles. Three clusters emerge: the two larger ones are named by their profiles, the third resembles both. (b)~Mean attention profiles of the repetition and novelty cluster over the seven query-symbol positions for each pattern. These means are the templates used to identify carriers across models. (c)~Label check on all \nheads{} heads of \model{}: each head's novelty-template loading against its attention share on the query's distinctive symbol (the symbol shown once, averaged over the five patterns that have one). Dashed vertical lines mark the mean of the repetition carriers, of all heads, and of the novelty carriers. (d)~For each of the 17 replication models, the number of its top-20 brain-scoring heads whose profile loads $\geq .50$ on the repetition template, on the novelty template, or on neither; the bottom row is the mean over models.}
    \label{fig:families}
\end{figure}

\paragraph{Brain-aligned heads reveal repetition and novelty profiles.}
To characterize the highest brain-scoring heads, we looked at their attention on the seven elements making up each sequence, averaged over the 25 trials of each pattern type, giving an $8 (patterns) \times 7 (elements)$ profile per head. Clustering the top 20 brain-scoring heads by these profiles yielded three clusters of seven, nine, and four heads (Figure~\ref{fig:families}a; Appendix~\ref{app:clusters}). Two clusters showed distinctive attention patterns shared across their constituent heads: the novelty cluster (nine heads) preferentially attended to the \emph{distinctive symbol}---the only symbol occurring once among the seven visible elements, a property present in five of the eight patterns---such as \texttt{B} in $\texttt{AAABAAA}\rightarrow\texttt{B}$, whereas the repetition cluster (seven heads) emphasises repeated-symbol positions, such as the second \texttt{A} (Figure~\ref{fig:families}b; Appendix~\ref{app:clusters}). It should be noted that these names are used here to describe attention profiles, not established computational functions. The novelty label is supported by a direct attention measure: across all heads, similarity to the novelty template increased with the share of attention on the query's distinctive symbol ($r=.73$; Figure~\ref{fig:families}c).
.

\paragraph{Template matching identifies carriers across models.}
We then averaged the profiles within each of these two clusters to obtain fixed templates and compared every head in the model with both templates. The comparison removes overall positional preferences (using a normalization and centering procedure further described in Appendix~\ref{app:pipeline-attention}) to emphasize how attention shifts across patterns. A head's \emph{loading} is defined as its Pearson correlation with a template, while a \emph{carrier} is a head whose loading exceeds the selection threshold of .50. Across heads, the two template loadings were strongly anticorrelated ($r=-0.83$), indicating contrasting attention organization rather than independent dimensions.

We then applied the frozen templates to every head of each cohort model. Averaged equally across models, repetition carriers accounted for \paperresult{carriers.top20.repetition}\% of the top 20 FRP heads versus \paperresult{carriers.all.repetition}\% of all heads, indicating modest overrepresentation. Novelty carriers were common but not overrepresented (\paperresult{carriers.top20.novelty}\% versus \paperresult{carriers.all.novelty}\%, respectively), and \paperresult{carriers.top20.neither}\% matched neither template (Appendix Figure~\ref{fig:replication-carrier-distributions}).

\paragraph{Independent clustering recovers related profiles.}
To test whether similar profiles emerge without imposing the templates, we independently clustered the attention profiles of each model's top 20 FRP-aligned heads, yielding \paperresult{clusters.top20.min}--\paperresult{clusters.top20.max} clusters per model (methods in Appendix~\ref{app:pipeline-clustering}). We then compared the three largest cluster means with the templates, finding a novelty-like profile ($r\geq.50$) in all \paperresult{clusters.matches.novelty} models and a repetition-like profile in \paperresult{clusters.matches.repetition}. Repetition- and novelty-like attention therefore generalized across models under independent clustering, despite differences in the number and composition of clusters.

\paragraph{Repetition-ranked removal is more damaging than novelty-ranked removal.}
Removing repetition heads in descending order of template similarity produced greater overall accuracy loss than random-head removal, following a similar trajectory to the concept-ranked ablation, albeit not as pronounced. Surprisingly, removing novelty heads had much less effect: accuracy dropped to \paperresult{exploration.ablation.novelty.at_800.accuracy}\% after removal of 800 heads (15.63\% of the model), compared with \paperresult{exploration.ablation.novelty.at_800.clean_accuracy}\% before ablation, while the random-control median fell below 50\%. The model therefore retained most of its task performance without this large set of novelty heads, whereas repetition-head removal was more disruptive.

The family ablations also generalized: in the cumulative sweeps (Figure~\ref{fig:replication-brain-ablation}), repetition-ranked removal was more damaging than random removal, with a peak excess loss of \paperresult{ablation.peak.repetition.loss} percentage points at \paperresult{ablation.peak.repetition.percent}\% of heads. Novelty-ranked removal, in contrast, never exceeded random removal by more than \paperresult{ablation.peak.novelty.loss} points at any tested fraction, and over most of the curve was \emph{less} damaging than removing the same number of random heads. The asymmetry between the two families observed in \model{} is therefore a property of the cohort, not of one model.

%% file: sections/05_families.tex
\section{What the attention families do and do not do}
\label{sec:pricing}
\label{sec:battery}

The analyses in this section probed the two families beyond the primary-task ablations of Section~\ref{sec:families} and were only run on \model{} and Pythia-1.4B. The latter was specifically used to study how these attention profiles and their effects on task performance evolved across training.

\paragraph{Repetition heads accompany the concept representation but do not supply it.}
In \model{}, two observations suggested that repetition heads might build the abstract-pattern representation carried by concept heads. First, across heads, repetition loading was positively related to concept score ($\rho=0.45$; Appendix Figure~\ref{fig:rep-cv}a), whereas novelty loading was negatively related ($\rho=-0.19$). Second, the two sets co-occurred in depth: the 130 heads with the highest repetition loading lay in layers 16--35 (median 28) and the 150 highest-scoring concept heads in layers 22--39 (median 33; Appendix Figure~\ref{fig:rep-cv}b). Did repetition-head removal therefore impair performance because it disrupted the representation of abstract patterns? It did not. Removing these 130 repetition heads lowered accuracy by 5.5 percentage points, but did not reduce cross-format concept RSA in later concept heads, and a classifier still identified the pattern from those heads' outputs with 100\% accuracy (Appendix~\ref{app:scaffold}; Figure~\ref{fig:rep-cv}c). Thus, the performance loss was not accompanied by a loss of the measured pattern information: repetition heads co-varied with the concept representation and occupied similar layers, but their removal did not eliminate the measured representation.

\paragraph{Novelty heads do not matter for other tasks, and did not earlier in training.}
The small effect of novelty-head removal might reflect specialization for another task. We therefore tested eight tasks spanning pattern completion, few-shot learning, prose, repeated-span verse, retrieval, coreference, and synthetic induction. Novelty-ranked removal was less damaging than the random median on seven tasks and fell within the random-control range on prose. In contrast, each task's own attribution ranking identified heads whose removal sharply impaired performance (Appendix Figure~\ref{fig:battery}). The contrast therefore extended beyond the original pattern completion task: the tested tasks were sensitive to head removal, but novelty-template similarity did not identify the heads on which they most depended.

We also tracked a fixed set of novelty-like heads across \paperresult{exploration.pythia.training.n_checkpoints} checkpoints of Pythia-1.4B \citep{biderman2023pythia}. Their attention increasingly resembled the final training profile. Measures of copying strength and attribution-based contribution to repeated-sequence prediction also increased, but remained below the reference thresholds used in that analysis. Separately, at the final checkpoint, removing the \paperresult{exploration.pythia.final.novelty32.n_heads_ablated} heads most similar to the Llama novelty template reduced pattern-task accuracy from \paperresult{exploration.pythia.final.novelty32.clean_accuracy}\% to \paperresult{exploration.pythia.final.novelty32.ablated_accuracy}\% (Appendix Figure~\ref{fig:pythia}). The growing prominence of this attention profile is therefore not matched by a growing role in the tested computations.

%% file: sections/06_gaze.tex
\section{Relationship to human gaze}
\label{sec:brain}

\begin{figure}[t]
\centering
\includegraphics[width=\linewidth]{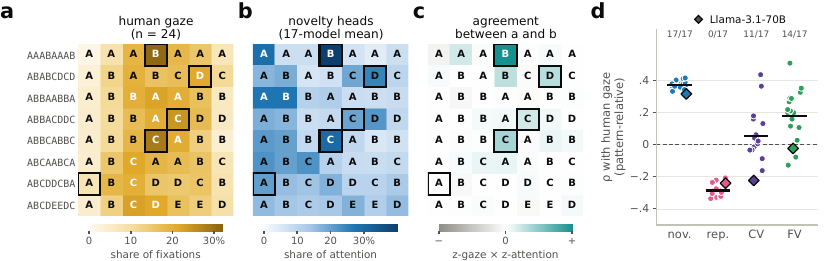}
\caption{\textbf{Human gaze and head-family attention.} (a)~Group fixation map of human participants over the seven visible query positions; each row is one pattern with its symbols overlaid, normalized to the share of that pattern's total fixation duration. (b)~Attention of the top 1\% novelty heads to the same positions, normalized per pattern and averaged over the 17 models. (c)~Cell-wise agreement between (a) and (b): both maps are centered per position, z-scored, and multiplied, so the cells average to the Pearson correlation between (a) and (b). Boxes in (a--c) mark the five patterns whose only unrepeated symbol is the correct continuation; 70\% of the agreement fell in these five cells and 95\% in their rows. The first-position case (\texttt{ABCDDCB → A}) drew neither gaze nor attention. (d)~Spearman correlation between the human map and the mean attention profile of each model's top 1\% of heads per family, after centering both per position (pattern-relative). One point per model (diamond: \model{}); bars are means over the 17 models, and counts give the models with $\rho>0$. Novelty heads correlated positively with gaze in all 17 models and repetition heads negatively in all 17.}
\label{fig:gaze}
\end{figure}

\paragraph{Novelty heads attend where humans look.} Because the human FRPs are time-locked to fixations on sequence elements, we additionally compared head attention with human gaze. The summed duration of fixations on each sequence icon was summarized on the same pattern-by-position grid as model attention (Figure~\ref{fig:gaze}a); Appendix~\ref{app:gaze-map} documents the map construction. For each of the 17 models we took the top 1\% of heads ranked by novelty loading, repetition loading, CV, FV, or FRP score, averaged their query-normalized $8\times7$ attention profiles, and correlated the result with the human map across the 56 cells. Novelty heads correlated positively with gaze in every model (mean $\rho=\paperresult{gaze.top1.novelty.relative}$, range $[\paperresult{gaze.top1.novelty.relative.min}, \paperresult{gaze.top1.novelty.relative.max}]$) and repetition heads negatively in every model (mean $\rho=\paperresult{gaze.top1.repetition.relative}$, range $[\paperresult{gaze.top1.repetition.relative.min}, \paperresult{gaze.top1.repetition.relative.max}]$), while CV and FV heads fell between them on average, with mean $\rho_{\mathrm{CV}}=\paperresult{gaze.top1.cv.relative}$ (range $[\paperresult{gaze.top1.cv.relative.min}, \paperresult{gaze.top1.cv.relative.max}]$) and mean $\rho_{\mathrm{FV}}=\paperresult{gaze.top1.fv.relative}$ (range $[\paperresult{gaze.top1.fv.relative.min}, \paperresult{gaze.top1.fv.relative.max}]$), respectively (Figure~\ref{fig:gaze}d).

\paragraph{What the correspondence reflects.} The task design confounds rarity with answer identity: in five of the eight patterns (\texttt{AAABAAAB}, \texttt{ABABCDCD}, \texttt{ABBACDDC}, \texttt{ABBCABBC}, \texttt{ABCDDCBA}) the only unrepeated symbol is also the correct continuation. The gaze correspondence was concentrated in these five (Figure~\ref{fig:gaze}c). Restricted to them, the pattern-relative correspondence of novelty heads averaged $.47$ across the 17 models, against $.12$ on the remaining three patterns. The repetition anticorrelation did not depend on this split ($-.34$ and $-.20$). Novelty heads and human gaze thus converged on the same cell, the single prior occurrence of the answer. Whether participants fixate it because it is the answer or because it is the odd one out cannot be decided from these patterns, since both accounts predict the same cell. What the ablations showed was that the model can attend to it without using it (Section~\ref{sec:families}). FV heads attended to the same symbol on the same five patterns ($.28$, versus $.00$ on the other three), so attention location alone does not separate idle heads from working ones.

%% file: sections/07_discussion.tex
\section{Discussion}
\label{sec:discussion}
We tested whether similarity to human frontal FRPs identified attention heads that mattered for abstract-pattern completion. Across the tested models, brain alignment was a weaker guide to performance sensitivity than patching scores, yet revealed shared repetition- and novelty-like attention profiles. 

\paragraph{Brain alignment is a weaker guide to performance-critical heads than causal scoring.}
Across \paperresult{cohort.models} models, brain alignment was weakly associated with patching effects, and brain-ranked ablation was substantially less damaging on average than FV-ranked ablation. The mean excess loss beyond random peaked at \paperresult{ablation.peak.fv.loss} percentage points after removing \paperresult{ablation.peak.fv.percent}\% of heads for FV, compared with \paperresult{ablation.peak.brain.loss} points at \paperresult{ablation.peak.brain.percent}\% for brain-ranked removal. Brain-ranked removal therefore reached less than one-third of FV's peak excess loss despite removing almost twice the fraction of heads; CV-ranked removal had an intermediate effect (Figure~\ref{fig:replication-brain-ablation}). Brain alignment was not a consistent proxy for abstract-pattern representation either: its association with concept scores was positive in Llama and base Qwen models but negative in instruction-tuned Qwen models and Phi-4. The clearest common finding was therefore the weak link between brain resemblance and performance sensitivity, rather than a universal link between brain resemblance and abstraction .

\paragraph{Repetition and novelty profiles recur across models.}
Clustering brain-aligned heads separately in each model recovered attention profiles resembling the repetition and novelty templates, even though the templates were not used to form the clusters. The distinction between attending to repeated and distinctive sequence elements therefore extended beyond the original Llama analysis. These profiles did not exhaust the variety of brain-aligned attention: some clusters matched neither template. Their recurrence identified shared patterns of attention across models, but does not by itself imply that heads with similar profiles perform the same computations.

\paragraph{Repetition heads support performance while pattern information remains available.}
In \model{}, repetition loading correlated with concept scores, and the two head sets occupied overlapping layers. Yet removing 130 strongly repetition-like heads reduced accuracy while preserving the measured pattern information in later concept heads. These heads therefore contributed to performance without being essential for that measured representation under the tested ablation. Repetition-like heads might also account for some of the effects of brain-ranked removal, but the two rankings select different, partly overlapping sets. Ablating repetition-like, novelty-like, and other heads within the same brain-selected population would test this explanation directly.

\paragraph{Novelty-like attention: human resemblance with limited task effects.}
Novelty-ranked removal was less damaging than random removal on average, despite the profile's recurrence across models. The additional task battery and final-model Pythia ablation also showed limited performance sensitivity to novelty-ranked heads. Novelty-like attention also followed pattern-dependent changes in human gaze after removing overall positional preferences. This correspondence was concentrated in patterns where the distinctive symbol is also the correct answer, so sensitivity to rarity cannot be separated from attention to answer-relevant information. One possible explanation is that novelty-like attention is a \emph{spandrel} \citep{gould1979spandrels}---a by-product of the model's architecture and training rather than a feature needed for these tasks. Softmax offers one reason this is possible: every head must distribute attention across the available tokens, even when its output has little influence on the answer. A recognizable attention profile can therefore coexist with a small task contribution. This does not explain why the profile favors distinctive elements, however, and our experiments do not establish its origin. Novelty-like heads resembled pattern-dependent changes in human gaze while contributing comparatively little to the tested tasks.

\paragraph{Relationship to other causal alignment studies.}
These findings need not conflict with evidence that training models to predict brain activity poorly impaired downstream performance \citep{merlin2026lose}, or that language-localized units were both causally important and more brain-aligned than random units \citep{alkhamissi2025llm}. Those studies differ in selection criteria, representational targets, interventions, and behavioral endpoints. Changing representations through training or selecting language-responsive units does not test the same proposition as ranking attention heads by task-specific brain alignment.

\paragraph{Limitations and future directions.}
The main comparison used eight sequence patterns, fixed three-shot prompts, and one group-average frontal-FRP target. Here, ``brain alignment'' refers specifically to similarity with this target, not to correspondence with human brain activity in general. It captures one aspect of the EEG data, and other electrode groups, time windows, signal features, or recording modalities such as fMRI could identify different model components. Testing these alternatives, alongside additional tasks and prompting strategies, would establish how broadly the observed dissociation generalizes. Additionally, the human RDM gives participants unequal weight because they contributed unequal amounts of recording data and contains only 28 dependent pattern comparisons. Additional patterns, held-out participants, and alternative RDM constructions would strengthen the assessment of brain alignment. Regarding the LLMs, testing them more extensively on other tasks and prompting strategies would assess the scope of the findings. In particular, we did not identify a substantial task-specific role for novelty-like heads in the tested settings, but this does not establish that they lack a causal role elsewhere. Testing their contribution to other computations and tasks is an important direction for future work. Finally, small zero-ablation effects may reflect overlapping contributions from other heads, and layer-matched controls would help separate head selection from network location.

\paragraph{Conclusion.}
In this study, we combined representational comparisons with head ablations to examine the relationship between brain alignment and causal importance. Across the tested models, brain alignment was a substantially weaker guide to performance-critical heads than patching scores. Brain-aligned heads nevertheless revealed recurring repetition- and novelty-like attention profiles with different relationships to performance: repetition-ranked removal was more disruptive, whereas novelty-like attention more closely resembled pattern-dependent changes in human gaze. Together, these findings show that resemblance to human representations and attention does not reliably identify the model components most important for producing a correct answer.

%% file: sections/08_appendix_A.tex
\section{Supporting methods and results}
\label{app:main-support}
This section follows the main-text sequence: human data and model evaluation, head scoring and ablation, attention families, extended interventions, and gaze comparisons.

\subsection{Human study and frontal-FRP target}
\label{app:rdm}
\label{app:human-study}

\paragraph{Participants and experimental setup.}
The source study \citep{pinier2025large} recruited 25 adults with normal or corrected-to-normal vision and no reported personal or family history of epilepsy. Five withdrew before completing all five sessions, their available sessions were retained in the original study. One thing to note is that these recruitment counts are distinct from the participants with usable FRP RDMs in the present reanalysis. The planned experiment comprised 400 unique trials over five sessions (80 per session, and 50 trials per pattern overall). Unlike the model prompts, the human trials did not contain three solved demonstrations, but participants performed three trials with feedback at the begining of their first session.

\begin{figure}[ht]
\centering
\includegraphics[width=\linewidth]{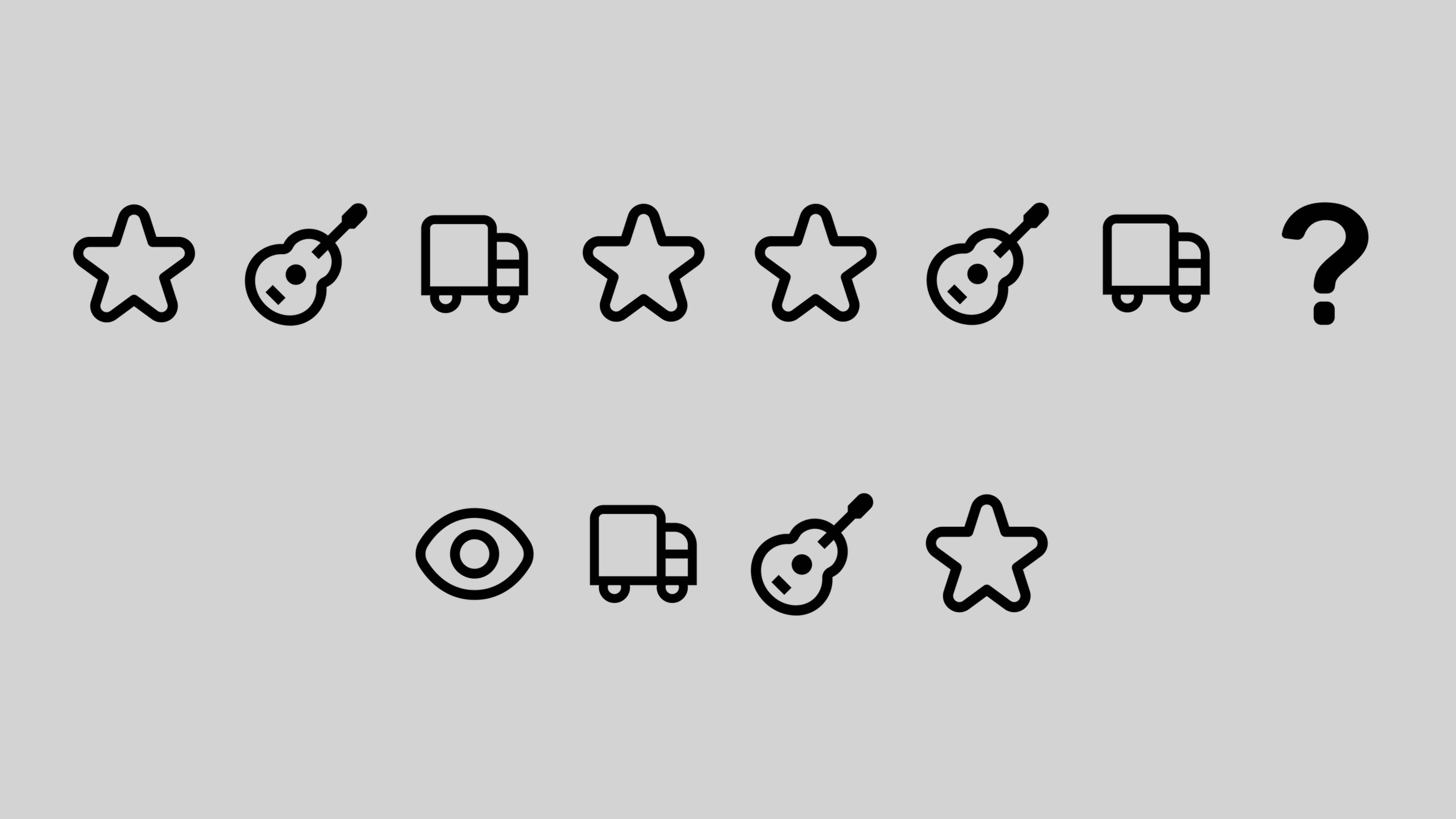}
\caption{\textbf{Example trial from the human task.} Seven icons instantiate an abstract pattern; the participant selects the missing eighth icon from four alternatives. Here the pattern is \texttt{ABCAABCA}, and the star completes it. The FRP analysis retains fixations within the eight sequence-position regions, including the question-mark position, but not the response-option regions. Adapted from \citet{pinier2025large}.}
\label{fig:human-trial}
\end{figure}

\paragraph{Recording and EEG preprocessing.}
The source study recorded 64-channel BioSemi EEG at 2,048 Hz and dominant-eye position with an EyeLink 1000 Plus at 2,000 Hz. Its documented MNE preprocessing marked bad channels, average-referenced the remaining channels, and applied zero-phase FIR notch filters at 50, 100, 150, 200, and 250 Hz. Extended Infomax ICA was fitted on a 1--100 Hz filtered copy; components not classified as brain by ICLabel were rejected, and the solution was applied to the notch-filtered data. A final 0.1--100 Hz bandpass, spherical-spline interpolation of bad channels, and average re-referencing completed preprocessing.

\paragraph{Fixation-related-potential construction.}
The frontal fixation-related-potential (FRP) signal was derived from the recordings reported by \citet{pinier2025large}:
for every retained fixation whose mean gaze position fell within one of the eight sequence-position bounding boxes (including the question mark), a 600-ms epoch of continuously preprocessed EEG beginning at fixation
onset was retained, without fixation-specific baseline correction. Fixations on answer options were excluded. Within a trial,
the retained epochs were averaged, so an FRP is a summary of self-paced
inspection of the sequence rather than the response to one experimentally
imposed event. Electrodes in the frontal region of interest were then averaged
at each time point to yield a trial-level temporal feature vector. This choice
is substantively important: the target can reflect visual inspection,
attention, and task difficulty as well as reasoning-related processing.
Consecutive fixation epochs can overlap when their onsets are less than 600 ms apart.

\begin{figure}[htbp]
\centering
\includegraphics[width=.65\linewidth]{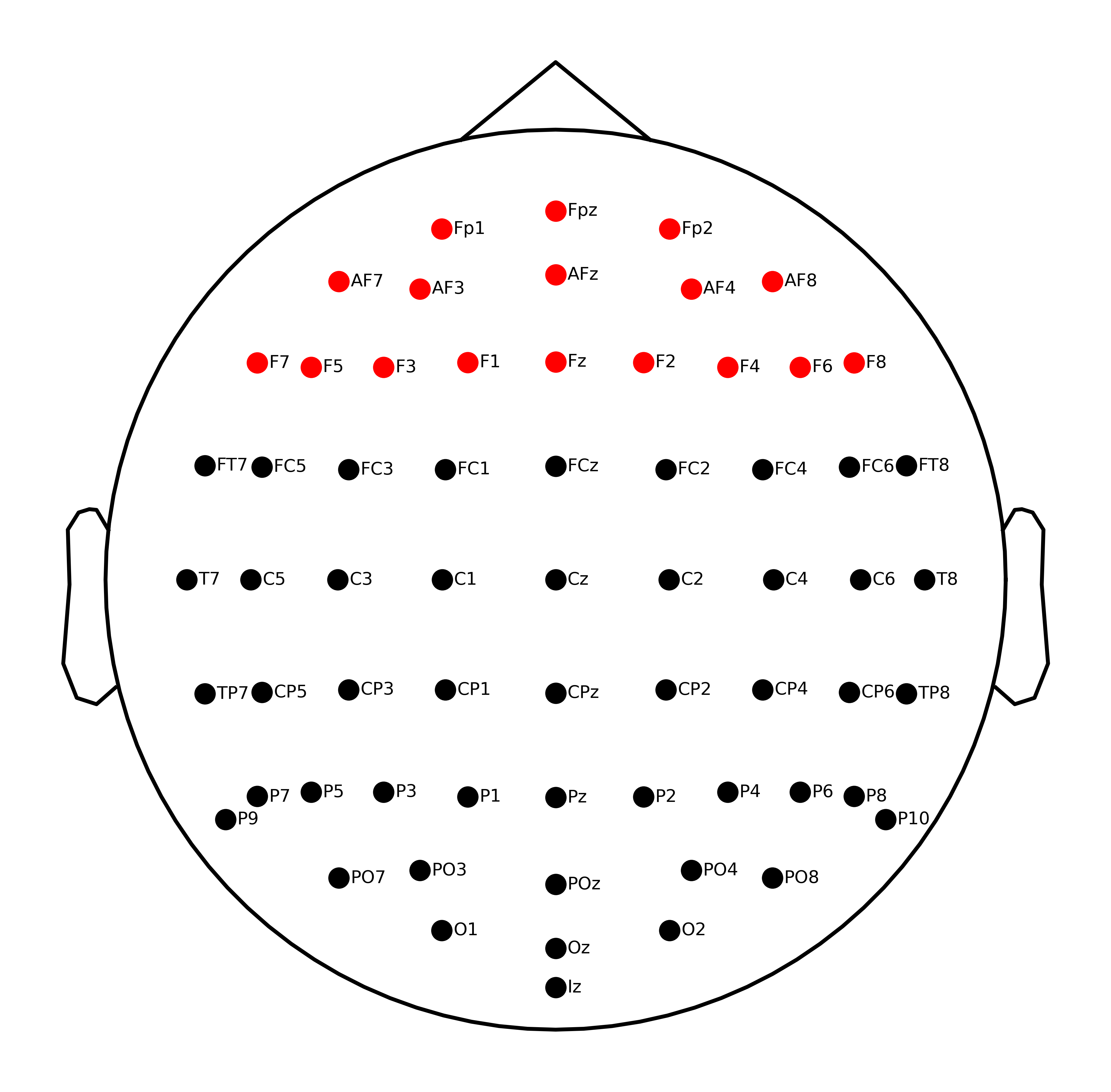}
\caption{\textbf{Frontal electrodes used for the human FRP target.} The 17 electrodes listed in the text are highlighted in red; other displayed montage electrodes are black. The scalp is viewed from above, with the nose at the top. Adapted from \citet{pinier2025large}.}
\label{fig:frontal-electrodes}
\end{figure}

\paragraph{Frontal electrode selection.}
The frontal region was an anatomical channel group, rather than electrodes selected for their correlation with model outputs. The analysis configuration specifies 17 channels (Figure~\ref{fig:frontal-electrodes}):
\texttt{Fp1, Fpz, Fp2, AF7, AF3, AFz, AF4, AF8, F7, F5, F3, F1, Fz, F2, F4, F6, F8}.
These frontopolar, anterior-frontal, and frontal electrodes were averaged with equal weight at each time point. This reduces each pattern's response to a temporal waveform; the resulting RDM therefore compares waveform shapes, not spatial scalp patterns.

\paragraph{The primary representational target.}
\label{app:frp-target}
Our primary target is the original, group-average frontal-FRP RDM used in the
initial \model{} discovery analysis. For each of the eight pattern classes,
group-averaged FRP data were summarized as a pattern-level temporal feature
vector; the dissimilarity of patterns $p$ and $q$ was their Pearson
correlation distance, $D_{pq}=1-r(\mathbf{e}_p,\mathbf{e}_q)$. This produces
one symmetric $8\times8$ RDM and hence 28 unique pattern-pair values. Thus,
the target retains the contribution of every eligible recording session in the
grand average before computing the RDM; participants with more retained data
can contribute more to this estimator.

\subsection{Model prompts and evaluation}
\label{app:task}

\paragraph{Prompt formats and evaluation conditions.}
The specification below describes the cross-model replication prompts. Table~\ref{tab:pipeline-stages} lists the surface alphabets, response formats, and trial counts used for each analysis stage.

\paragraph{Abstract patterns and item generation.} The model task contains eight length-eight abstract patterns, \texttt{AAABAAAB}, \texttt{ABABCDCD}, \texttt{ABBAABBA}, \texttt{ABBACDDC}, \texttt{ABBCABBC}, \texttt{ABCAABCA}, \texttt{ABCDDCBA}, and \texttt{ABCDEEDC}. The eighth element is withheld from the query. For every item, abstract letters are mapped without replacement to surface tokens, yielding seven displayed sequence elements and one correct continuation. The generator creates 25 independently sampled items for every pattern, alphabet, and response format, with random seed 42. The three alphabets are punctuation-like symbols, German words, and English words; the two response formats are open-ended completion and four-alternative multiple choice. Distractors are sampled first from other values used in the mapping and then, if needed, from unused values in the same alphabet. The correct answer and three distractors are shuffled deterministically.

\paragraph{Three-shot prompt.} Every prompt contains three solved demonstrations followed by one unsolved query. The demonstrations instantiate the \emph{same abstract pattern} as the query but use fresh, non-overlapping surface tokens. No natural-language instruction, system message, or additional introductory example is present. An open-ended demonstration is rendered as a space-separated seven-element sequence followed on the next line by \texttt{Answer: $x$}; a multiple-choice demonstration inserts an \texttt{Options:} block containing four shuffled surface-token alternatives before its answer. Demonstrations are separated by blank lines. The query has the same structure but ends at the literal boundary \texttt{Answer:}, with no trailing space and no supplied answer. The scored continuation is the correct eighth surface token. The current cross-model replication used this prompt with raw text (no chat template), even for instruction-tuned models. This holds prompt construction fixed; tokenizer-specific word pools can change literal lexical items. Chat-template and alternative-instruction experiments were separate prompt studies and are not combined with the results reported here.

\paragraph{Tokenization and continuation target.} Word pools are filtered separately for each tokenizer: an eligible word must be represented by one token both in isolation and when preceded by a space. The preflight stage checks every content pool, option, and correct completion; it also verifies the exact prompt--continuation boundary by tokenizing the prompt alone and the prompt plus candidate continuation. A trial is accepted only when the latter adds exactly one token without changing the prefix token sequence.

\begin{table}[t]
\centering
\small

\resizebox{\linewidth}{!}{%
\begin{tabular}{lll}
\toprule
Stage & Model trials & Used to construct \\
\midrule
Head RSA & 1,200 trials across six conditions & CV and control scores \\
Brain RSA & 200 English MC trials (RSA subset) & human-RDM scores \\
Attention scan & 200 English MC trials & fingerprints and templates \\
Attribution patching & 200 English open-ended trials & FV/AIE ranking \\
Independent clustering & saved fingerprints & unlabeled attention groups \\
Set construction & saved scores & full rankings and intervention grid \\
Zero-ablation sweep & 200 English MC trials & causal trajectories and metrics \\
\bottomrule
\end{tabular}}
\caption{Stage-specific trial conditions in the replication pipeline. MC denotes multiple choice. The six conditions cross English words, German words, and symbols with MC and open-ended completion, with 25 items per pattern and 200 trials per condition. Brain scores reuse the English MC subset of the head-RSA dataset; the attention scan and ablation sweep use a separately generated English MC set.}
\label{tab:pipeline-stages}
\end{table}

\subsubsection{Evaluation conditions and performance}
\label{app:replication-performance}
Table~\ref{tab:replication-clean-performance} lists the included models and their clean-task performance under the common prompt protocol specified in Appendix~\ref{app:task}.

\paragraph{Formats and scoring.}
Performance is evaluated on the trial conditions specified in Table~\ref{tab:pipeline-stages}. Our primary measure is \emph{exact next-token accuracy}: whether the highest-probability token across the full vocabulary is the contextually resolved correct answer token. This is not accuracy restricted to the four options, nor an evaluation of extended reasoning or generated explanations.

\begin{table}[t]
\centering
\small

\begin{tabular}{lrr}
\toprule
Model family and size & Base & Instruct / post-trained \\
\midrule
Llama-3.2-3B & \paperresult{clean.llama_3.2_3b} & \paperresult{clean.llama_3.2_3b_instruct} \\
Llama-3.1-8B & \paperresult{clean.llama_3.1_8b} & \paperresult{clean.llama_3.1_8b_instruct} \\
Llama-3.1-70B & \paperresult{clean.llama_3.1_70b} & \paperresult{clean.llama_3.1_70b_instruct} \\
Llama-3.3-70B & --- & \paperresult{clean.llama_3.3_70b_instruct} \\
\midrule
Qwen2.5-3B & \paperresult{clean.qwen2.5_3b} & \paperresult{clean.qwen2.5_3b_instruct} \\
Qwen2.5-14B & \paperresult{clean.qwen2.5_14b} & \paperresult{clean.qwen2.5_14b_instruct} \\
Qwen2.5-32B & \paperresult{clean.qwen2.5_32b} & \paperresult{clean.qwen2.5_32b_instruct} \\
Qwen2.5-72B & \paperresult{clean.qwen2.5_72b} & \paperresult{clean.qwen2.5_72b_instruct} \\
\midrule
Phi-4 (14B) & --- & \paperresult{clean.phi_4} \\
DeepSeek-R1-Distill-Llama-70B & --- & \paperresult{clean.deepseek_r1_distill_llama_70b} \\
\bottomrule
\end{tabular}
\caption{\textbf{Clean English multiple-choice performance.} Exact next-token accuracy (\%) on 200 trials per model under raw, three-shot prompting. Each populated cell is one of the \paperresult{cohort.models} models. Dashes indicate no included model, not missing trials. Phi-4 and DeepSeek-R1-Distill-Llama are post-trained models; their placement in the right column does not imply identical training procedures.}
\label{tab:replication-clean-performance}
\end{table}

\paragraph{Scale, family, and instruction tuning.}
Performance generally improved with size within Llama and Qwen2.5-Instruct, but size alone did not determine success: Qwen2.5 base peaked at 32B (\paperresult{clean.qwen2.5_32b}\%) rather than 72B (\paperresult{clean.qwen2.5_72b}\%), and Phi-4 reached \paperresult{clean.phi_4}\% with 14B parameters. Instruction tuning improved the larger Qwen models and Llama-3.1-70B, but not the 3B Qwen or 3B/8B Llama pairs under this raw prompt. DeepSeek's \paperresult{clean.deepseek_r1_distill_llama_70b}\% measures immediate completion, not the benefit of allowing a reasoning continuation. These are descriptive comparisons under a fixed prompting protocol, not isolated causal effects of parameter count or post-training.

\paragraph{Sensitivity to format.}
Six-format accuracy evaluations are available only for a subset and are kept separate from the \paperresult{cohort.models}-model sweep. In standalone raw-prompt evaluations of Llama-3.1-8B and 70B base, English open-ended accuracy was \paperresult{inference.llama8.open_ended}\% and \paperresult{inference.llama70.open_ended}\%, versus \paperresult{inference.llama8.multiple_choice}\% and \paperresult{inference.llama70.multiple_choice}\% for English multiple choice; German-word performance showed the same direction, while symbol conditions yielded only \paperresult{inference.symbols.min}--\paperresult{inference.symbols.max}\% exact next-token accuracy. These evaluations regenerate the full six-condition set, so their English multiple-choice items need not match the sweep items. Prompt formatting also mattered: in a paired Llama-3.3-70B-Instruct evaluation, adding the chat template raised English multiple-choice accuracy from \paperresult{inference.llama33.words_eng.raw}\% to \paperresult{inference.llama33.words_eng.chat}\% and symbol multiple-choice accuracy from \paperresult{inference.llama33.symbols.raw}\% to \paperresult{inference.llama33.symbols.chat}\%. Thus, low performance in the primary condition does not establish an absence of abstract-pattern competence. Subsequent causal comparisons use each model's own clean baseline under the shared English multiple-choice protocol.

\subsection{Brain, concept, and function-vector scores}

\subsubsection{Head outputs, concept score, and control RSA}
\label{app:pipeline-rsa}

For each prompt, the pipeline extracts the input to the attention output projection at the final prompt position. This tensor is the concatenation of the individual attention-head output vectors before they are mixed by the layer's output projection; it is reshaped into one vector per layer and head. For each head, all trial vectors in the head-RSA dataset (Table~\ref{tab:pipeline-stages}) are $\ell_2$-normalized and their pairwise cosine similarities are computed.

The \emph{concept score}, used to rank candidate concept-vector (CV) heads, is Spearman's $\rho$ between these head-output similarities and a binary design matrix whose entries indicate whether two items instantiate the same abstract pattern. The format and alphabet controls are computed identically, using same-format and same-alphabet design matrices. These are continuous rankings over every head, not intrinsically thresholded head families; a ``top-$k$ CV set'' means the first $k$ heads in the descending concept-score ordering. The use of all alphabets and both response formats broadens the criterion beyond one condition, but does not prove format invariance or causal use.

\paragraph{Head RDMs and brain scores.}
\label{app:brain-score}
For each head, outputs from the English multiple-choice subset (Table~\ref{tab:pipeline-stages}) were $\ell_2$-normalized trial by trial and averaged within each pattern. Writing the mean for pattern $p$ as $\bar{\mathbf{v}}_p$, the between-pattern RDM entries were $D_{pq}=1-\bar{\mathbf{v}}_p^{\mathsf T}\bar{\mathbf{v}}_q$. The pattern means were not normalized again: their dot product equals the mean cosine similarity across all trial pairs from the two patterns, which is how the implementation computed these entries. A head's brain score was Spearman's $\rho$ between the 28 upper-triangular entries of this RDM and the corresponding entries of the group-average frontal-FRP RDM (Appendix~\ref{app:frp-target}).

\subsubsection{Function-vector score by attribution patching}
\label{app:pipeline-ap}

The function-vector (FV) ranking is estimated on 200 English-word, open-ended trials. A clean forward pass first averages each head's final-position output vector over the complete clean dataset. The corrupted dataset preserves vocabulary, prompt scaffold, response format, and target token, but independently randomizes the diagnostic prefix of every demonstration and query. Corruptions are rejected if they preserve the clean canonical pattern; demonstration answers are replaced by plausible values from the corrupted sequence. This destroys recoverable abstract-pattern information while avoiding a wholesale change of surface vocabulary.

On the corrupted prompts, the pipeline computes the gradient of the original correct answer token's probability with respect to each head's final-position output. The first-order attribution for a head and trial is the inner product of this gradient with the displacement from the corrupted activation to that head's clean mean activation. Averaging over trials gives the estimated average indirect effect (AIE). Heads are ranked in descending mean AIE to form the FV ordering. Model parameters are frozen; gradients are retained through activations by rooting autograd at the input embedding, and a loss scale is divided out before scores are saved. Runs containing non-finite probabilities or attributions are rejected. This score is an efficient first-order approximation to mean activation patching, not the cumulative zero-ablation effect used later.

\subsubsection{Concept and function vectors on the pattern task}
\label{app:cvfv}

The CV and FV rankings introduced in Section~\ref{sec:dissociation} define comparison sets at explicitly stated sizes, rather than categorical head types. The replication calculations are detailed in Appendices~\ref{app:pipeline-rsa} and~\ref{app:pipeline-ap}; the latter uses attribution rather than full activation patching. Their overlap is calculated by exact layer--head identity, and their causal consequences are compared on the common cumulative zero-ablation grid. This design separates three questions that are otherwise easy to conflate: whether a head's output geometry represents pattern identity, whether restoring that head is predicted to increase correct-answer probability, and whether removing a growing set of such heads actually changes behavior.

\subsubsection{Attribution patching validation}
\label{app:ap}
In the exploratory analysis of \model{}, first-order attribution patching was compared with full single-head mean activation patching on the same trials and prompt configuration. Across heads, effect estimates had Pearson $r=.983$ and Spearman $\rho=.60$; the top-20, top-50, and top-150 head sets overlapped by 95\%, 90\%, and 91\%, respectively. The strong agreement in effect magnitude and high-effect selections supports attribution patching as an efficient FV-selection method here. The lower overall rank correlation means that the complete rankings are not interchangeable, and this single-model validation does not establish equivalent accuracy on every replication model.

Attribution patching and cumulative zero ablation also test different interventions: the former approximates replacing one corrupted activation with its clean mean, whereas the latter removes sets of heads from clean prompts. We therefore use the sweep, rather than the patching score alone, to measure the behavioral consequences of the FV selection.

\subsection{Cumulative ablation and cross-model score comparisons}

\subsubsection{Rankings and cumulative zero ablation}
\label{app:pipeline-ablation}

The set-construction stage creates five complete descending head orderings: FV (mean AIE, or full activation patching when explicitly available), brain (the selected primary human RDM), CV (concept RSA), repetition loading, and novelty loading. Non-finite values are placed last. Five seed-0 uniform random permutations of all heads form the standard descriptive reference band. The cumulative grid is $1,2,3,5,8,13,20,32,50,80,125,200,320,500,800,1250,2000,3200,5120$, truncated to values below the model's total head count and followed by that total. Thus each point removes a progressively larger set from the top of one fixed ranking; the random reference is five nested random trajectories, not a fresh sample at every head count.

Clean and ablated performance are evaluated on the same sweep trial set (Table~\ref{tab:pipeline-stages}). For every ordering and head count, the selected heads' slices in the input to each attention output projection are set to zero at \emph{every token position}. Slices are computed from the actual output-projection input width and number of attention heads, which also accommodates grouped-query architectures. The model is otherwise unchanged. All rankings necessarily converge when every head is removed, providing an implementation check. The primary behavioral outcome is exact next-token accuracy over the full vocabulary: the model's largest final-position logit must be the contextually resolved correct continuation token. The sweep output additionally stores the correct-token logit and probability, log-normalizer, top-two logits, full-vocabulary argmax, KL divergence from the clean next-token distribution, all four option logits, option-restricted prediction and accuracy, item pattern, and clean baselines. These arrays support per-pattern and alternative-metric analyses without rerunning the model.

The five standard random trajectories matched the ablated head count but not the layer distribution.

\subsubsection{Cross-model summaries}
\label{app:replication-causal}
\begin{figure}[t]
    \centering
    \includegraphics[width=.93\linewidth,height=.64\textheight,keepaspectratio]{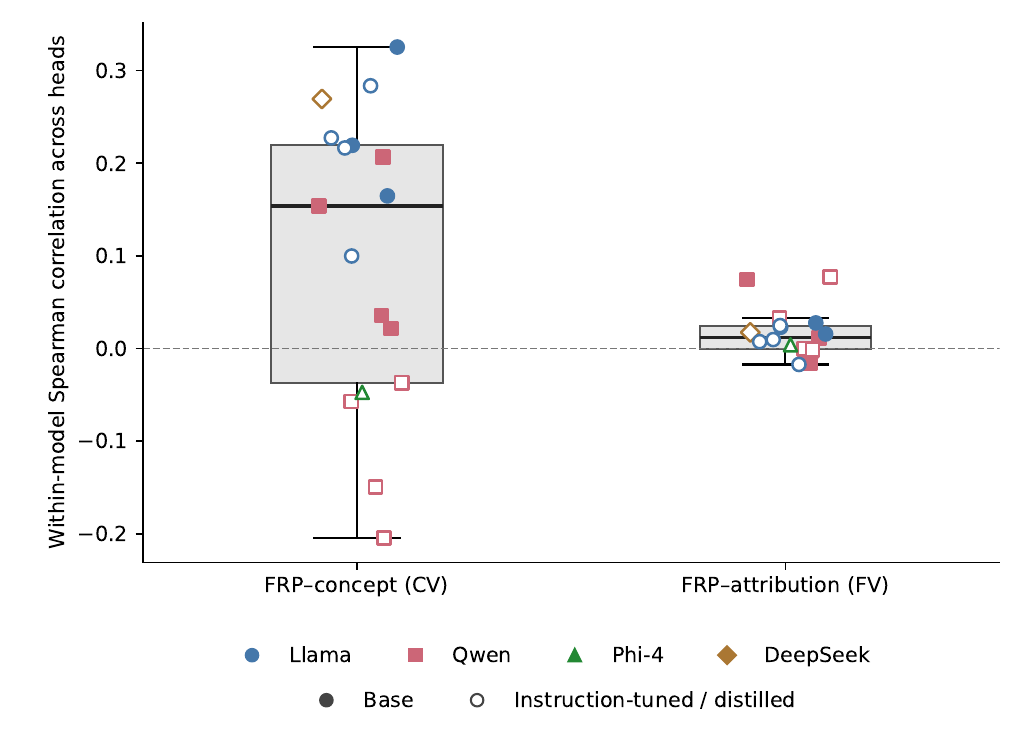}
    \caption{\textbf{Brain alignment and head-selection scores.} Each point represents one model's Spearman correlation across heads between frontal-FRP RSA and concept-RSA (CV) or attribution-patching (FV) scores; colors identify model families. Boxes show medians and interquartile ranges, with whiskers extending to the most extreme observations within 1.5 interquartile ranges. All model points are displayed. Undefined head scores are excluded pairwise. The \paperresult{cohort.models} models are not independent samples of model families; these descriptive correlations are not participant-level inference.}
    \label{fig:replication-brain-correlations}
\end{figure}

\begin{figure}[ht]
    \centering
    \begin{minipage}[c]{0.47\linewidth}
    \caption{\textbf{Brain score against concept, patching and alphabet scores.} Per cohort model, the Spearman correlation across heads between brain score and concept (CV), attribution-patching (FV) or alphabet (control) score. Circles: Llama, including the DeepSeek distill; squares: Qwen2.5; triangle: Phi-4. Open markers are tuned models, the outlined circle is \model{}, and bars are family means. The brain--concept correlation was positive in every Llama and base Qwen model and negative in every tuned Qwen model and Phi-4; the other correlations stayed near zero.}
    \label{fig:brain-corr}
    \end{minipage}\hfill
    \begin{minipage}[c]{0.49\linewidth}
    \includegraphics[width=\linewidth]{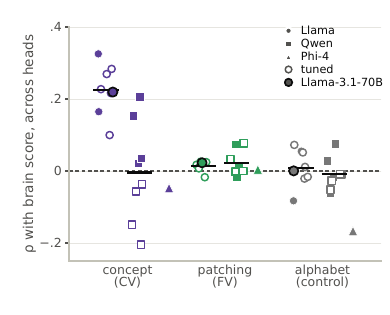}
    \end{minipage}
\end{figure}

Head-score associations were Spearman correlations across heads within each model, excluding undefined scores pairwise (Figures~\ref{fig:replication-brain-correlations} and~\ref{fig:brain-corr}).

For the cumulative sweeps, head counts were expressed as percentages of each model's total heads and trajectories were linearly interpolated between tested counts. We subtracted each model's mean random-control accuracy and averaged the resulting differences equally across models. Negative values in Figure~\ref{fig:replication-brain-ablation} therefore indicate greater impairment than random removal. Reported peak excess losses are maxima of the equal-model mean loss curves, not averages of model-specific maxima. Model-bootstrap intervals summarize variation within this cohort; related models are not independent samples of model families.

\subsection{Attention profiles, templates, and brain-selected clusters}

\subsubsection{All-head attention scan and frozen family loadings}
\label{app:pipeline-attention}

The attention scan saves every head's attention row at the final prompt position for the trials specified in Table~\ref{tab:pipeline-stages}. Token spans are reconstructed from the rendered prompt, and the seven positions occupied by the unsolved query sequence are selected. Attention rows are first averaged over the 25 trials of each pattern. Each head's seven query-position values in this pattern mean are then divided by their sum, producing an $8\times7$ matrix. In the shared fingerprint implementation, averaging precedes query normalization. Finally, each position's mean across the eight patterns is removed. Flattening this matrix gives a 56-dimensional, pattern-centered attention fingerprint. Heads that place effectively no attention on the seven-symbol span have undefined template loadings and are sorted last in template rankings.

The frozen templates are the two discovery-cluster means described in Section~\ref{sec:families}; they are not refitted in the replication. For every head in every model, a family loading is the Pearson correlation between its pattern-centered fingerprint and the corresponding frozen template. Cross-model descriptive analyses use a common carrier cutoff of .50. Cumulative template ablations do not stop at the carrier cutoff: they rank \emph{all} heads by loading and remove increasingly long prefixes.

\subsubsection{Independent attention-profile clustering}
\label{app:clusters}
\label{app:pipeline-clustering}
For the top-20 analyses, pairwise distances between flattened attention profiles (Appendix~\ref{app:pipeline-attention}) were defined as $d=1-r$, where $r$ is Pearson correlation. Average-linkage hierarchical clustering merged the groups with the smallest mean pairwise distance; the dendrogram was cut at distance .50. This cut does not require every pair of heads within a cluster to have $r\geq.50$. The three largest cluster means were compared with the frozen templates after clustering, using $r\geq.50$ to identify a match.

\subsubsection{Template carriers, independent clusters, and membership agreement}
\label{app:replication-clusters}
\begin{figure}[t]
\centering
\includegraphics[width=\linewidth]{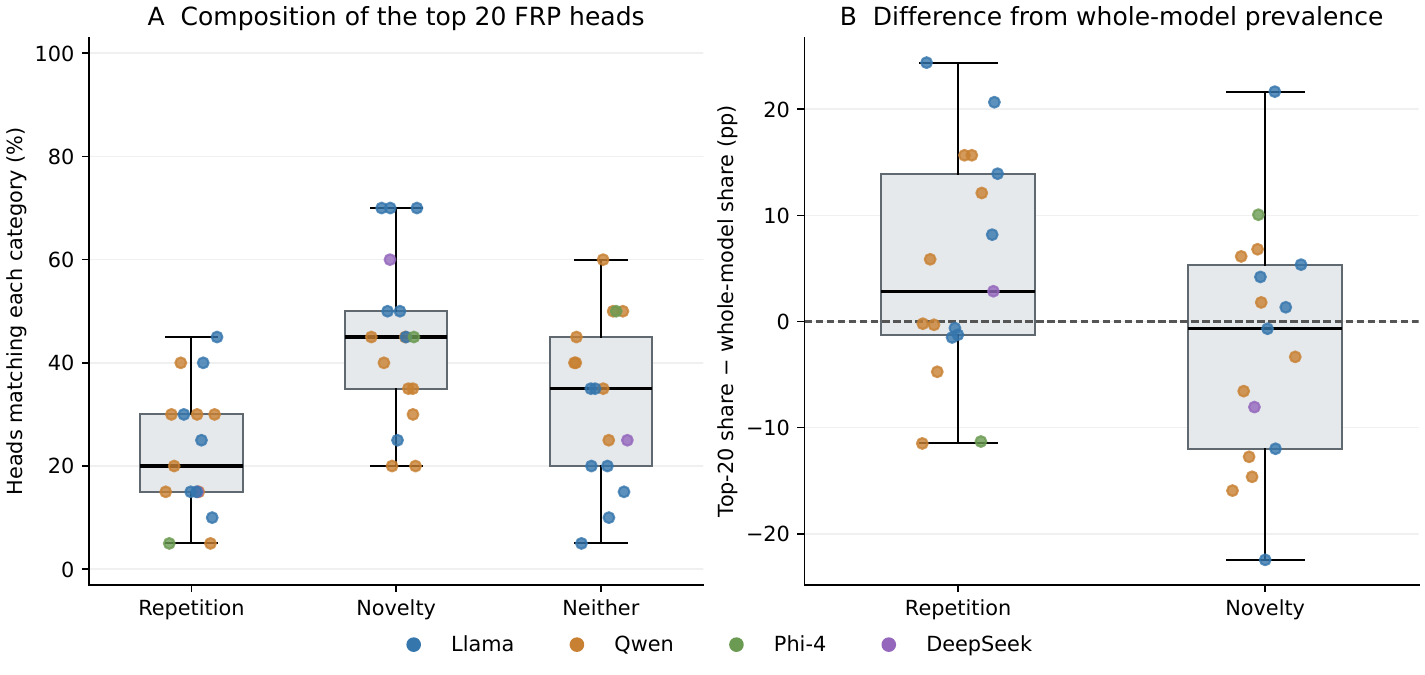}
\caption{\textbf{Template carriers among the top 20 FRP heads.} Each point is one of the \paperresult{cohort.models} models; colors identify model families. Boxes show medians and interquartile ranges, with whiskers extending to the most extreme observations within 1.5 interquartile ranges. (A) Percentage of the top 20 heads matching repetition, novelty, or neither template at $r\geq.50$. (B) Within-model difference between top-20 carrier share and its prevalence among all model heads; zero indicates equal prevalence. Positive values indicate overrepresentation, not a significance test.}
\label{fig:replication-carrier-distributions}
\end{figure}

\paragraph{Carrier prevalence.} For each model, we computed carrier proportions among all heads and among its top 20 FRP heads using the cutoff in Appendix~\ref{app:pipeline-attention}. Overrepresentation was the within-model difference between these proportions; cohort summaries weighted models equally (Figure~\ref{fig:replication-carrier-distributions}).

\paragraph{Membership agreement.}
Each template-carrier set was restricted to the same top-20 candidate pool and matched to the independent cluster with maximum Jaccard overlap (intersection divided by union). Mean overlap was \paperresult{overlap.repetition} for repetition and \paperresult{overlap.novelty} for novelty; exact set equality occurred in \paperresult{overlap.exact.repetition} and \paperresult{overlap.exact.novelty} of \paperresult{cohort.models} models, respectively. These post-hoc best matches measure set agreement, not independent predictive validation.

\begin{figure}[t]
    \centering
    \includegraphics[width=\linewidth]{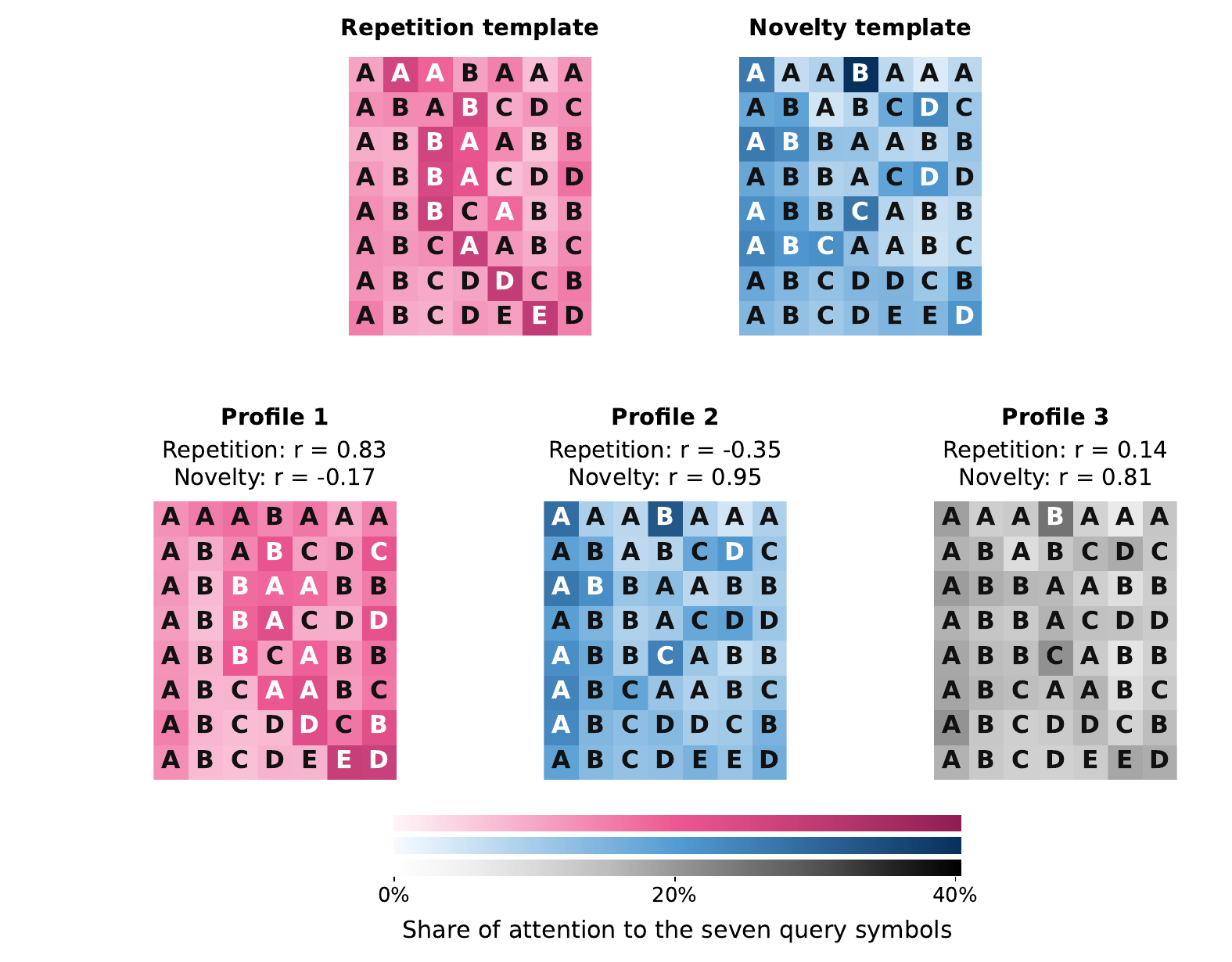}
    \caption{\textbf{Templates and independently discovered cluster profiles.} Top row: the original repetition and novelty templates. Bottom row: equal-model mean profiles of the three largest clusters among each model's top 20 FRP heads. Rows are patterns; columns are query positions, with the sequence symbols overlaid. Color intensity shows within-query attention share on a common scale; pink and blue link the first two aligned profiles to their display references, not to a categorical template assignment. For display, profiles were assigned one-to-one to the original cluster profiles by maximizing total Pearson similarity, without changing memberships. Title correlations compare each displayed mean profile with the indicated template using Pearson correlation after centering each position across patterns; they are not averages of model-wise correlations. This alignment is not independent evidence of recurrence or of three clusters per model. Post-hoc template comparisons found novelty-like matches in \paperresult{clusters.matches.novelty}/\paperresult{cohort.models} models and repetition-like matches in \paperresult{clusters.matches.repetition}/\paperresult{cohort.models} at $r\geq.50$.}
    \label{fig:replication-cluster-profiles}
\end{figure}

Figure~\ref{fig:replication-cluster-profiles} shows the cluster profiles and their post-hoc template correspondence.

\subsection{Extended tests of attention-family function}
\label{app:extensions}
These extensions examine representation, task performance, training, and pruning beyond the standardized replication sweep.

\subsubsection{Repetition-head ablation and the concept representation}
\label{app:scaffold}
\begin{figure}[htbp]
    \centering
    \includegraphics[width=\linewidth]{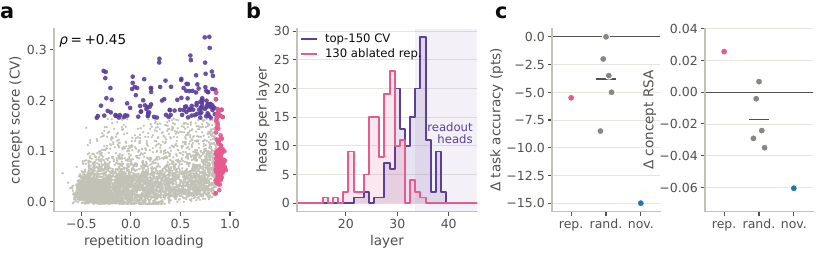}
    \caption{\textbf{Repetition-head removal reduced accuracy but preserved measured pattern information} (\model{}). (a)~Repetition-template loading against concept score for all heads (Spearman $\rho$). Pink: the 130 ablated repetition heads; purple: the 150 highest-scoring concept (CV) heads; gray: all other heads. (b)~Layer distribution of the same two sets. The shaded band marks the readout heads, the concept heads in layers 34 and above. (c)~Change from the unablated model after zero-ablating the 130 repetition heads (rep.), five damage-matched random 25-head sets (rand.; bar: mean), or 130 layer-matched novelty carriers (nov.). Left: multiple-choice task accuracy. Right: cross-format concept RSA over the readout heads. A pattern classifier on the readout heads stayed at 100\% accuracy in every condition.}
    \label{fig:rep-cv}
\end{figure}

This analysis tested whether the repetition heads build the concept representation, given that the two were correlated across heads and lay in overlapping layers (Figure~\ref{fig:rep-cv}a,b).

\paragraph{Head sets.} The \emph{repetition set} contains the 130 heads with the highest repetition-template loading (all loadings $\geq.85$), spanning layers 16--35 (median 28). The \emph{concept set} contains the 150 heads with the highest concept score, spanning layers 22--39 (median 33); nine heads belonged to both sets. All layers are 1-indexed.

\paragraph{Controls.} Two controls accompany the repetition set. The \emph{novelty control} contains 130 novelty carriers matched to the repetition set's layer profile; 103 heads required a nearest-layer substitute (median shift 3 layers, maximum 5). The \emph{damage-matched random} sets are drawn from heads that carry neither template, with the repetition set's layer proportions. Their size was chosen by bisection so that their mean loss in correct-answer log-probability matches the repetition set's loss. This yielded sets of 25 heads, of which we drew five. The match was imperfect at this small effect size: the random sets cost more log-probability than the repetition set ($0.090$ vs.\ $0.059$ nats) but less accuracy ($3.8$ vs.\ $5.5$ points).

\paragraph{Readouts.} Each set is zero-ablated in the English multiple-choice and open-ended formats (200 trials each), and we record the outputs of the concept heads at the final position. The readout heads are the concept-set heads in layers 34 and above that are not themselves ablated (73 heads; 72 for two random draws), so the measurement lies downstream of most of the repetition set. \emph{Cross-format concept RSA} is computed per head as the Spearman correlation between its multiple-choice and open-ended item RDMs ($1-$cosine similarity), averaged over readout heads. It is compared with the unablated value on the same heads. The \emph{pattern classifier} is a multinomial logistic regression over the concatenated readout-head outputs, trained on unablated multiple-choice trials (5-fold cross-validated accuracy $100\%$; permutation null $12.6\%$). It is then frozen and applied to the ablated outputs.

\paragraph{Results.} Removing the repetition set lowered accuracy from $70.5\%$ to $65.0\%$, while concept RSA rose slightly from $0.401$ to $0.427$ (Figure~\ref{fig:rep-cv}c). The damage-matched random sets lowered accuracy by a mean of $3.8$ points (range $[0, 8.5]$) and reduced concept RSA by $0.017$ on average (SD $0.016$). The novelty control lowered accuracy by $15$ points and reduced concept RSA by $0.060$. The classifier identified the pattern on $100\%$ of trials in every condition. Removing the repetition heads therefore cost task accuracy without degrading the measured concept representation, and degraded that representation less than removing random or novelty heads.

\subsubsection{Battery details}
\label{app:battery}
\begin{figure}[h]
\centering
\includegraphics[width=\linewidth]{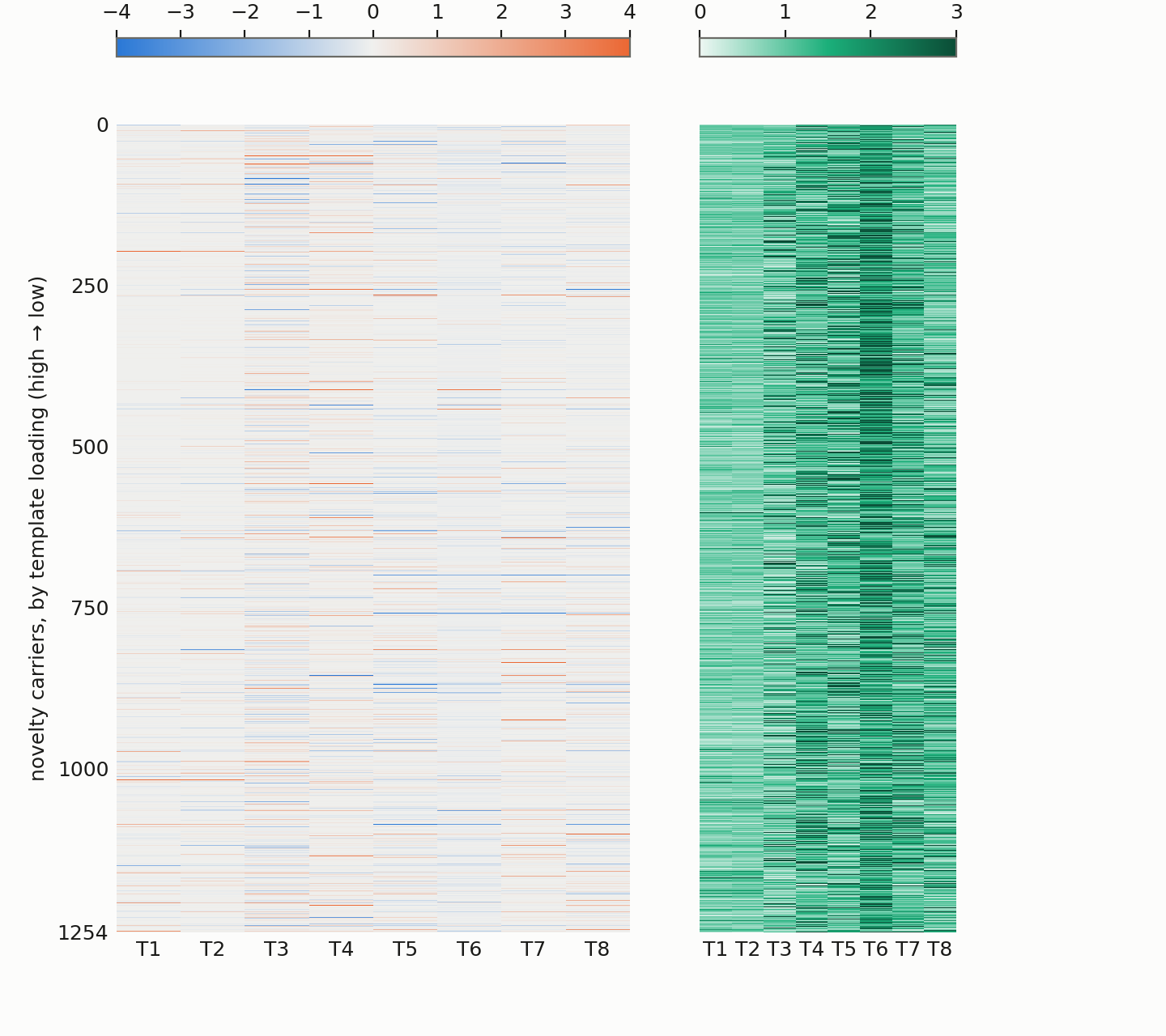}

\caption{Attribution price of every novelty carrier on every task (z against layer-matched non-carriers) beside the gate column.}
\end{figure}

The earlier description identifies pattern completion in two formats (T1--T2), few-shot in-context learning (T3), prose (T4), repeated-span verse (T5), needle retrieval (T6), coreference cloze (T7), and synthetic induction (T8).

\begin{figure}[htbp]
\centering
\includegraphics[width=\linewidth]{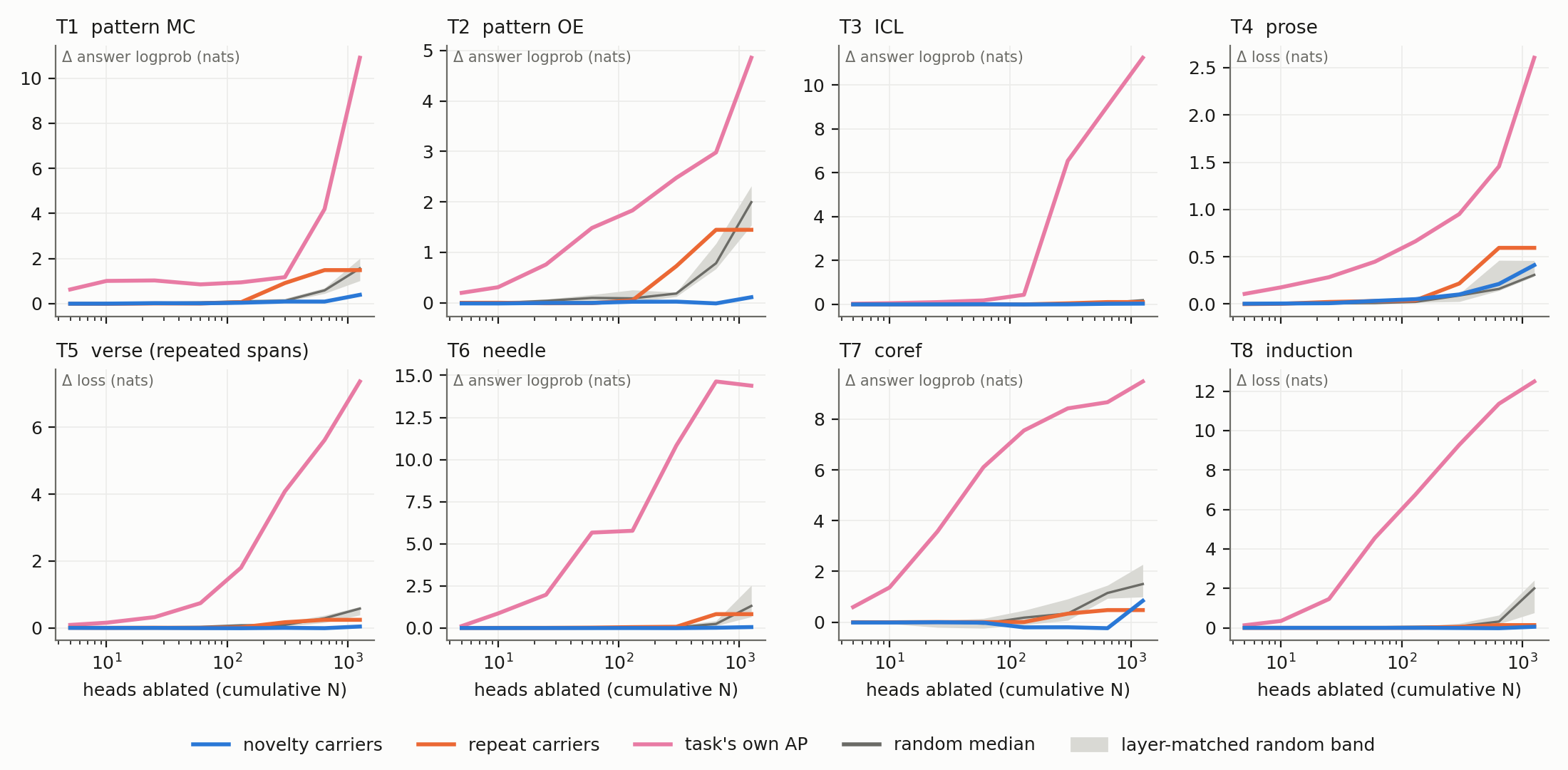}
\caption{Exploratory eight-task cumulative-ablation trajectories.}
\label{fig:battery}
\end{figure}

\subsubsection{Pythia trajectories}
\label{app:pythia}

\begin{figure}[ht]
    \centering
    \includegraphics[width=\linewidth]{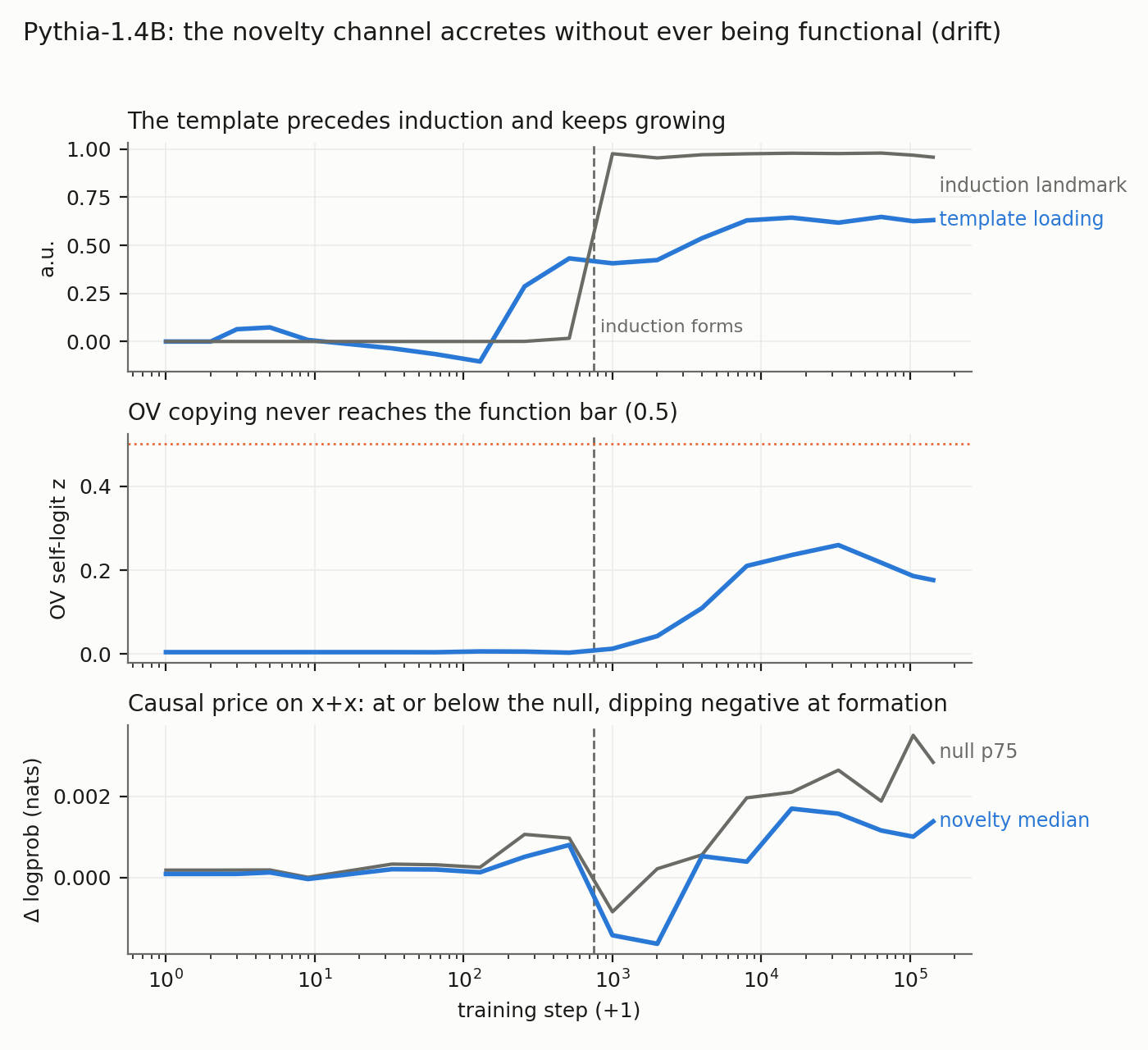}
    
    \caption{Template loading, OV copying and causal price of the future novelty population across Pythia-1.4B training.}
\end{figure}

\begin{figure}[htbp]
    \centering
    \includegraphics[width=\linewidth]{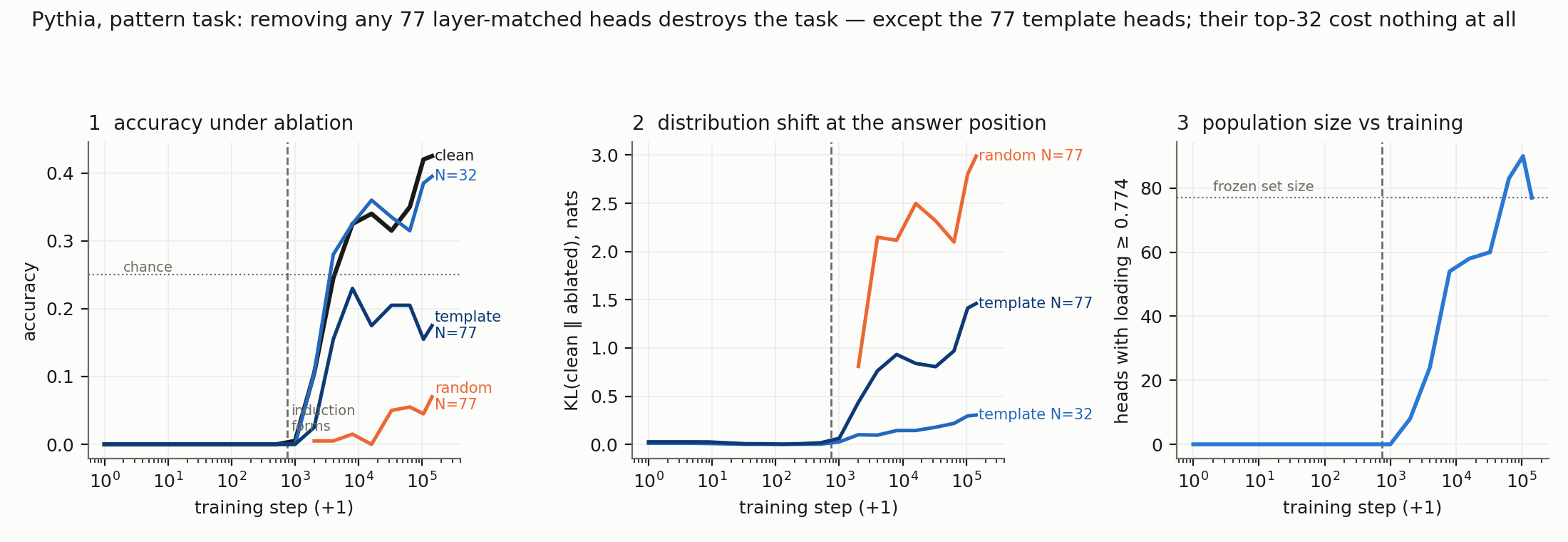}
    \caption{Exploratory Pythia pattern-task ablation, KL and template-population trajectories. These concern the tested interventions, not universal functionality.}
    \label{fig:pythia}
\end{figure}

\subsection{Human gaze and model attention}

\paragraph{Gaze-map construction.}
\label{app:gaze-map}
The human gaze heatmaps were constructed by assigning fixations to sequence-item positions and summing the durations of all fixations on each icon. Trials are grouped by abstract pattern within each participant, pooling sessions before averaging; participant maps are then averaged with equal participant weight. The seven visible query positions are retained for comparison with the model profiles, excluding the eighth sequence position and answer options. Unfixated positions receive zero fixation duration within otherwise observed trials; trials without usable sequence fixations are marked missing. The original export includes eight sequence positions, whereas the subsequent comparison uses seven-position files.

\subsubsection{Exploratory gaze comparison}
\label{app:discovery-gaze}
The exploratory gaze comparison yielded template correlations $.55$/$-.39$ and participant means $.41$/$-.28$ for novelty/repetition. These are distinct from the top-1\% replication comparisons shown in Figure~\ref{fig:gaze}.
The human map construction is documented in Appendix~\ref{app:gaze-map}.

\paragraph{Gaze geometry and positional attention are different comparisons.}
Using the human gaze maps described in Appendix~\ref{app:gaze-map}, we performed two distinct comparisons. Comparing head-family attention RDMs with the human group gaze RDM over the seven visible positions yielded mean model-wise RSA of $\paperresult{gaze.rdm.novelty}$ for novelty carriers and $\paperresult{gaze.rdm.repetition}$ for repetition carriers. This tests which \emph{patterns} have similar attentional organization, not whether models and humans attend to identical positions. Direct positional comparison instead used each model's top 1\% of heads ranked separately by novelty, repetition, CV, FV, or FRP score---not the thresholded carrier sets above. Their query-normalized attention was averaged into an $8\times7$ profile and correlated with the corresponding human group gaze profile across 56 cells.

Raw positional correspondence was weakly negative for novelty and repetition (mean model-wise $\rho=\paperresult{gaze.top1.novelty.direct}$ and $\paperresult{gaze.top1.repetition.direct}$). After separately subtracting each position's across-pattern mean from the model and human profiles, novelty correspondence became positive ($\paperresult{gaze.top1.novelty.relative}$) and repetition correspondence negative ($\paperresult{gaze.top1.repetition.relative}$), with these respective signs in all \paperresult{cohort.models} models. CV, FV, and brain-selected groups showed smaller positive pattern-relative means ($\paperresult{gaze.top1.cv.relative}$, $\paperresult{gaze.top1.fv.relative}$, and $\paperresult{gaze.top1.top.relative}$; Figure~\ref{fig:gaze}d). Restricting selection to the top 0.5\% retained the novelty/repetition pattern ($\paperresult{gaze.top0.5.novelty.relative}$/$\paperresult{gaze.top0.5.repetition.relative}$). Thus, novelty-like heads resembled \emph{pattern-dependent deviations} in gaze allocation despite differing in overall positional preference; repetition-like heads showed an opposing relative relationship. These normalized analyses do not measure how much total attention a model assigns to the sequence rather than prompt scaffolding, and mean correlations across models must not be confused with the correlation of a grand-average profile.